\documentclass[conference]{IEEEtran}
\IEEEoverridecommandlockouts
\usepackage{cite}
\usepackage{amsmath,amssymb,amsfonts}
\usepackage{algorithmic}
\usepackage{graphicx}
\usepackage{textcomp}
\usepackage{xcolor}
\usepackage{booktabs}
\usepackage{multirow}
\usepackage{amsmath}
\usepackage{enumitem}

\usepackage[figuresleft]{rotating}
\def\BibTeX{{\rm B\kern-.05em{\sc i\kern-.025em b}\kern-.08em
    T\kern-.1667em\lower.7ex\hbox{E}\kern-.125emX}}
\begin{document}

\title{Image Data Augmentation Approaches: A 
 Comprehensive Survey and Future directions\\
{\footnotesize \textsuperscript{*}Corresponding author(s)}
\thanks{This research was supported by Science Foundation Ireland under grant numbers 18/CRT/6223 (SFI Centre for Research Training in Artificial intelligence), SFI/12/RC/2289/$P\_2$ (Insight SFI Research Centre for Data Analytics), 13/RC/2094/$P\_2$ (Lero SFI Centre for Software) and 13/RC/2106/$P\_2$ (ADAPT SFI Research Centre for AI-Driven Digital Content Technology). For the purpose of Open Access, the author has applied a CC BY public copyright licence to any Author Accepted Manuscript version arising from this submission.}
}

\author{\IEEEauthorblockN{1\textsuperscript{st} Teerath Kumar\textsuperscript{*}}
\IEEEauthorblockA{CRT-AI, ADAPT Research Centre \\ School of Computing,\\ Dublin City University, Ireland; \\ teerath.menghwar2@mail.dcu.ie}
\and

\IEEEauthorblockN{2\textsuperscript{nd} Alessandra Mileo}
\IEEEauthorblockA{INSIGHT \& I-Form Research Centre  \\ School of Computing, \\ Dublin City University, Ireland;\\ alessandra.mileo@dcu.ie}
\and
\IEEEauthorblockN{3\textsuperscript{rd} Rob Brennan}
\IEEEauthorblockA{ADAPT Research Centre \\School of Computer Science, \\ University College Dublin, Ireland; \\ rob.brennan@adaptcentre.ie}
\and

\IEEEauthorblockN{4\textsuperscript{th}  Malika Bendechache}
\IEEEauthorblockA{ADAPT \& Lero Research Centres, \\ School
of Computer Science, \\ University of Galway, Ireland; \\ malika.bendechache@universityofgalway.ie}
}

\maketitle

\begin{abstract}
Deep learning algorithms have demonstrated remarkable performance in various computer vision tasks, however, limited labeled data can lead to overfitting problems, hindering the network's performance on unseen data. To address this issue, various generalization techniques have been proposed, including dropout, normalization, and advanced data augmentation. Among these techniques, image data augmentation - which increases the dataset size by incorporating sample diversity - has received significant attention in recent times. In this survey, we focus on advanced image data augmentation techniques. We provide an overview of data augmentation, present a novel and comprehensive taxonomy of the reviewed data augmentation techniques, and discuss their strengths and limitations. Furthermore, we provide comprehensive results of the impact of data augmentation on three popular computer vision tasks: image classification, object detection, and semantic segmentation. For results reproducibility, the available codes of all data augmentation techniques have been compiled. Finally, we discuss the challenges and difficulties, as well as possible future directions for the research community. This survey provides several benefits: i) readers will gain a deeper understanding of how data augmentation can help address overfitting problems, ii) researchers will save time searching for comparison results, iii) the codes for the data augmentation techniques are available for result reproducibility, and iv) the discussion of future work will spark interest in the research community.


\end{abstract}


\section{Introduction \& motivation}
Deep learning models have gained popularity and achieved tremendous progress in computer vision (CV) tasks such as image classification~\cite{he2016deep,krizhevsky2012imagenet,simonyan2014very,kumar2021binary,kumar2021class,chandio2022precise,khan2022introducing,roy2022wildect}, object detection~\cite{he2017mask,girshick2014rich}, , image segmentation~\cite{liu2019recent,lempitsky2009image,kuruvilla2016review,liew2017regional} and medical imaging~\cite{tataei2022glioma,ranjbarzadeh2022breast, ranjbarzadeh2022brain,ranjbarzadeh2023me,baseri2022investigation,ranjbarzadeh2022mrfe}. This advancement has been propelled by various deep neural network architectures, powerful computation resources and extensive availability of data~\cite{shorten2019survey}. Convolutional Neural Networks (CNNs) have demonstrated remarkable performance in CV tasks among all deep learning models. CNNs learn different features of an image by applying the convolution operation with the input image and kernel. The initial layers of CNN learn low-level features (e.g. edges, lines) while deeper layers learn more structured and complex features. The success of CNNs has stimulated the interest to use them for CV tasks. In addition to CNNs, Vision Transformers (ViT)~\cite{dosovitskiy2020image} are also gaining popularity and have been widely used in deep learning for CV tasks.

However, these algorithms are data-intensive and often suffer from the overfitting problem~\cite{russakovsky2015imagenet} - where the model performs well on training data but poorly on test data (unseen data). The issue is exacerbated when large amounts of data are not available, which can occur due to privacy concerns or the need for time-consuming and expensive human labeling tasks~\cite{shorten2019survey,kumar2021binary}. Despite the existence of large datasets such as ImageNet~\cite{deng2009imagenet}, overfitting remains a challenge because the standard training process only learns the important regions, but fails to learn less important features that are necessary for generalization~\cite{yun2019cutmix}. Moreover, adversarial attacks~\cite{hendrycks2021natural,zhao2017generating,madry2017towards} pose a threat to the accuracy of CNNs, where small, invisible perturbations added to the input image can fool the network and cause it to fail to identify the correct features in an image.

To address these challenges, data augmentation is often applied, not just in CV tasks, but also in a range of domains such as audio~\cite{turab2022investigating, park2020search,kumar2020intra,chandio2021audd,aleem2022random,ko2015audio,nanni2020data,singh2023understanding} and text~\cite{feng2020genaug,liu2020survey,shorten2021text,bayer2021survey}. This survey will specifically focus on the CV domain.

\begin{figure*}[!htbp]
    \centering
    \includegraphics[width=0.8\textwidth, height=5.5cm]{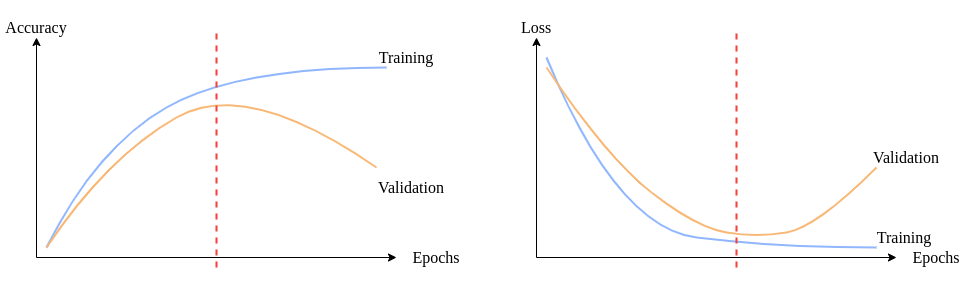}
    \caption[Caption]{Overfitting problem: On the left side, overfitting is explained in terms of accuracy, after the inflation point (red dotted line), the training accuracy is increasing but validation accuracy is decreasing. On the right side, alternatively in terms of loss, training loss is decreasing but validation loss is increasing after the red dotted line.  The figure is taken from the source \footnotemark{https://www.baeldung.com/cs/ml-underfitting-overfitting}}
    \label{fig:overfitting}
\end{figure*}

Regularization is an effective method for generalizing Convolutional Neural Network (CNN) models from both architectural and data perspectives. Various forms of regularization have been developed, including Data Augmentation~\cite{zhong2020random}, Dropout~\cite{srivastava2014dropout}, Batch Normalization~\cite{ioffe2015batch}, Transfer Learning~\cite{weiss2016survey,shao2014transfer}, and Pre-training~\cite{erhan2010does}. Among these, Image Data Augmentation~\cite{zhong2020random} has proven to be a useful form of regularization in several studies~\cite{krizhevsky2017imagenet,zhong2020random,simonyan2014very}. This technique expands the dataset by altering the sample's appearance or flavor~\cite{zhong2020random} to provide a more diverse range of views. However, performing Image Data Augmentation directly on image data can increase the risk of overfitting and biases, making it both important and challenging, as discussed further in Section~\ref{discussion}. 
 \newline  
 Generally, Image Data Augmentation addresses two primary problems in CNN models. The first problem is a shortage of data or limited data, which can result in overfitting. Image Data Augmentation provides a solution by feeding the model with various scenarios of an image, making the model more generalized and allowing for the extraction of more information from the original dataset. The second problem relates to labeling, where the original dataset has a label for each sample. Augmenting the sample preserves the label of the original sample and assigns it to the augmented sample.

Numerous surveys have been conducted on the topic of image data augmentation. For example,  Wang et al. explored and compared several traditional data augmentation techniques in their work ~\cite{perez2017effectiveness}, but this study was limited to image classification tasks only. In another study, Wang et al. reviewed the available data augmentation approaches for face recognition ~\cite{wang2020survey}. Khosla et al. briefly discussed warping and oversampling-based data augmentation approaches in their work ~\cite{khosla2020enhancing}. However, the authors did not provide a comprehensive taxonomy or a thorough evaluation of the techniques they discussed. Shorten et al. presented a comprehensive survey on image data augmentation in their work ~\cite{shorten2019survey}. The authors proposed a novel taxonomy, discussed future directions, and addressed the challenges associated with data augmentation. However, the survey lacked an evaluation of image data augmentation for various computer vision tasks. Additionally, as the study is three years old, it may not include the latest state-of-the-art augmentation methods such as cutmix and grid mask.  . Recently, Yang et al. conducted a survey on data augmentation in computer  vision tasks~\cite{yang2022image}. However, their study only covered a few data augmentation methods and did not provide any code compilation for result reproducibility.   Another study by Xu proposed a novel taxonomy for image data augmentations~\cite{xu2022comprehensive}, but did not evaluate the techniques discussed. This paper presents an extended taxonomy for data augmentation and reviews state-of-the-art techniques. The source code used in this study is available for result reproducibility. It should be noted that this survey does not cover data augmentations based on generative adversarial networks (GANs) due to out of the scope of this paper. But we redirect the reader to ~\cite{su2020survey,yue2022survey} for more details about GAN-based data augmentations.

The followings are our contributions: 

\begin{itemize}
\item A comprehensive image data augmentation taxonomy is presented.
\item An extensive survey of state-of-the-art data augmentation techniques, complete with visual examples, is provided.
\item The performance of state-of-the-art data augmentation techniques is evaluated and compared for several computer vision tasks.
\item The challenges of data augmentation are highlighted and future directions are identified.
\item The available codes for data augmentations, following the proposed taxonomy, are compiled for result reproducibility and made available at~\footnote{https://github.com/kmr2017/Advanced-Data-augmentation-codes}.
\end{itemize}

The above contributions provide the following benefits: 
\begin{itemize}
    \item  A better understanding of data augmentation working mechanism to fix the overfitting problem. 
    \item Our comprehensive analysis and comparison between the existing data augmentation techniques will save researchers time searching this field.  
    \item   Facilitates result reproducibility by providing the source code for the different data augmentation techniques investigated.
    \item  Future work will spark interest in the research community.

\end{itemize}

\begin{sidewaysfigure*}[!htbp]
    \centering
    \includegraphics[width=1.1\textwidth, height=16cm]{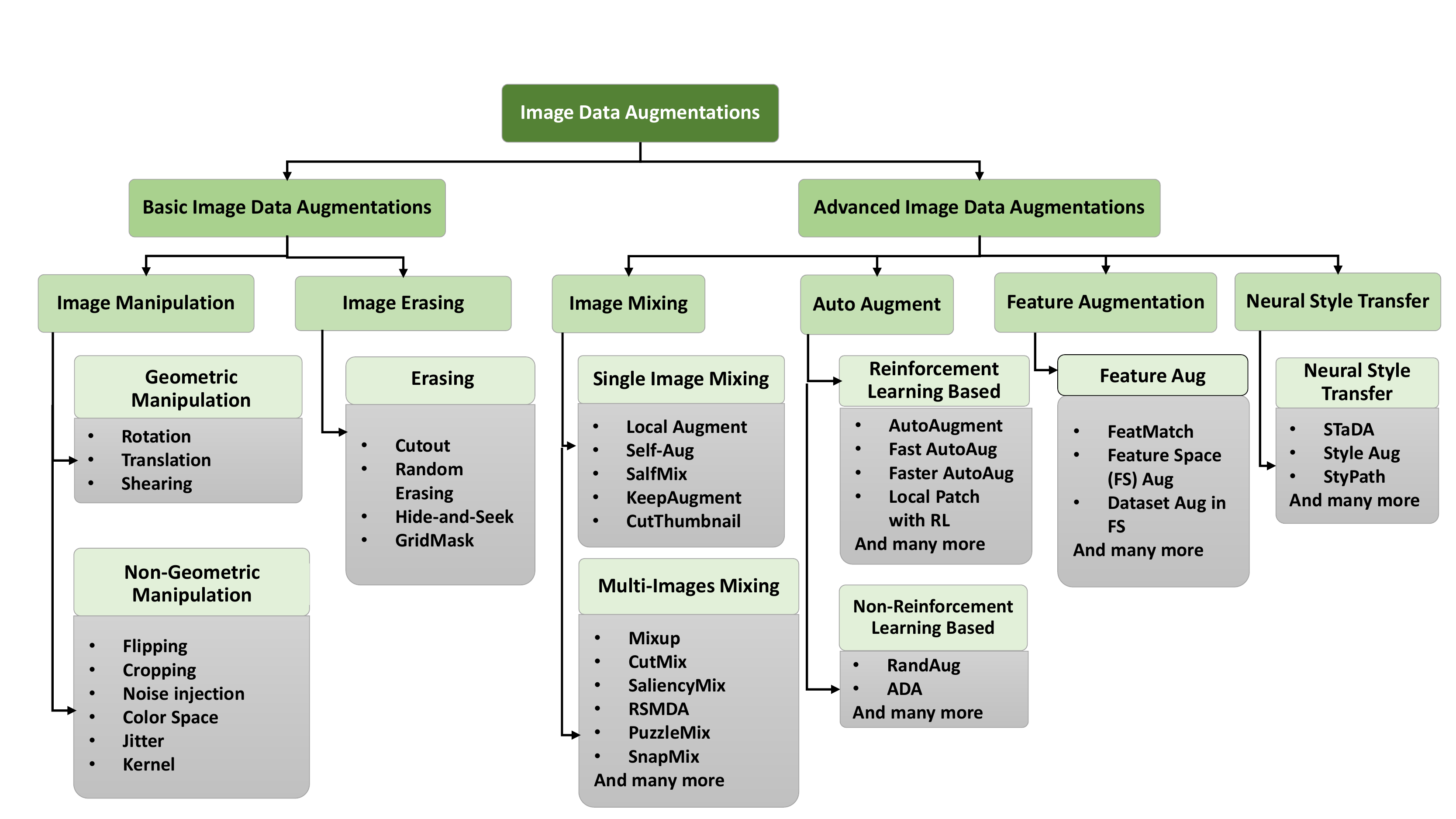}
    \caption{Image data augmentation taxonomy. Note: All image data augmentation names are not added in this taxonomy due space limit. However, all relevant and remaining image data augmentations are discussed as per taxonomy. The remaining sub-type of categories are discussed in the text.  }
    \label{fig:multimodality_Arch}
\end{sidewaysfigure*}

\section{Taxonomy and background}
The proposed taxonomy, presented in Figure~\ref{fig:multimodality_Arch}, classifies data augmentation into two main branches: Basic and Advanced data augmentations. The former encompasses fundamental techniques for data augmentation, while the latter encompasses more complex techniques. The specifics of each data augmentation method are thoroughly discussed in subsequent sections.
\subsection{\textbf{Basic Image Data Augmentations}}
This section describes basic image data augmentation methods and their classifications. They are classified as below:
\begin{itemize}
    \item \textbf{Image Manipulation}
    \begin{itemize}
        \item \textbf{\textit{Geometric Manipulation}}
        \item \textbf{\textit{Non-Geometric Manipulation}} 
    \end{itemize}
    \item \textbf{Image Erasing}
    \begin{itemize}
        \item \textbf{\textit{Erasing}}
    \end{itemize}
\end{itemize}
\subsubsection{\textbf{Image Manipulation}}
Image manipulation refers to the changes made in an image with respect to its position or color. Positional manipulation is made by adjusting the position of the pixels while color manipulations are made by altering the pixel values of the image. Image manipulation is further divided into two main categories. Each of them is discussed below: 

\textbf{Geometric Data Augmentation:} Geometric data augmentation encompasses modifications to the geometric attributes of an image, including its position, orientation, and aspect ratio. This technique involves transforming the arrangement of pixels within an image through a variety of techniques such as rotation, translation, and shearing. Figure~\ref{fig:geometric} illustrates the most commonly employed geometric augmentations. These methods are widely used in the domain of computer vision to diversify the training data and improve the resilience of models to diverse transformations. The utilization of geometric data augmentation has become a critical component in the development of robust computer vision algorithms. Each of the geometric data augmentations is discussed below: 
\begin{enumerate}[label=(\roman*)]
\item {\textbf{Rotation:}} Rotation data augmentation involves rotating an image by a specified angle within the range of 0 to 360 degrees. The precise degree of rotation is a hyperparameter that requires careful consideration based on the nature and characteristics of the dataset. For instance, in the MNIST~\cite{deng2012mnist} dataset, rotating all digits by 180 degrees, transforming a right-rotated 6 results into a 9, would not be a meaningful transformation. Therefore, a thorough understanding of the dataset is necessary to determine the optimal degree of rotation and achieve the best results.

\begin{figure}[htbp]
{\includegraphics[width=0.5\textwidth]{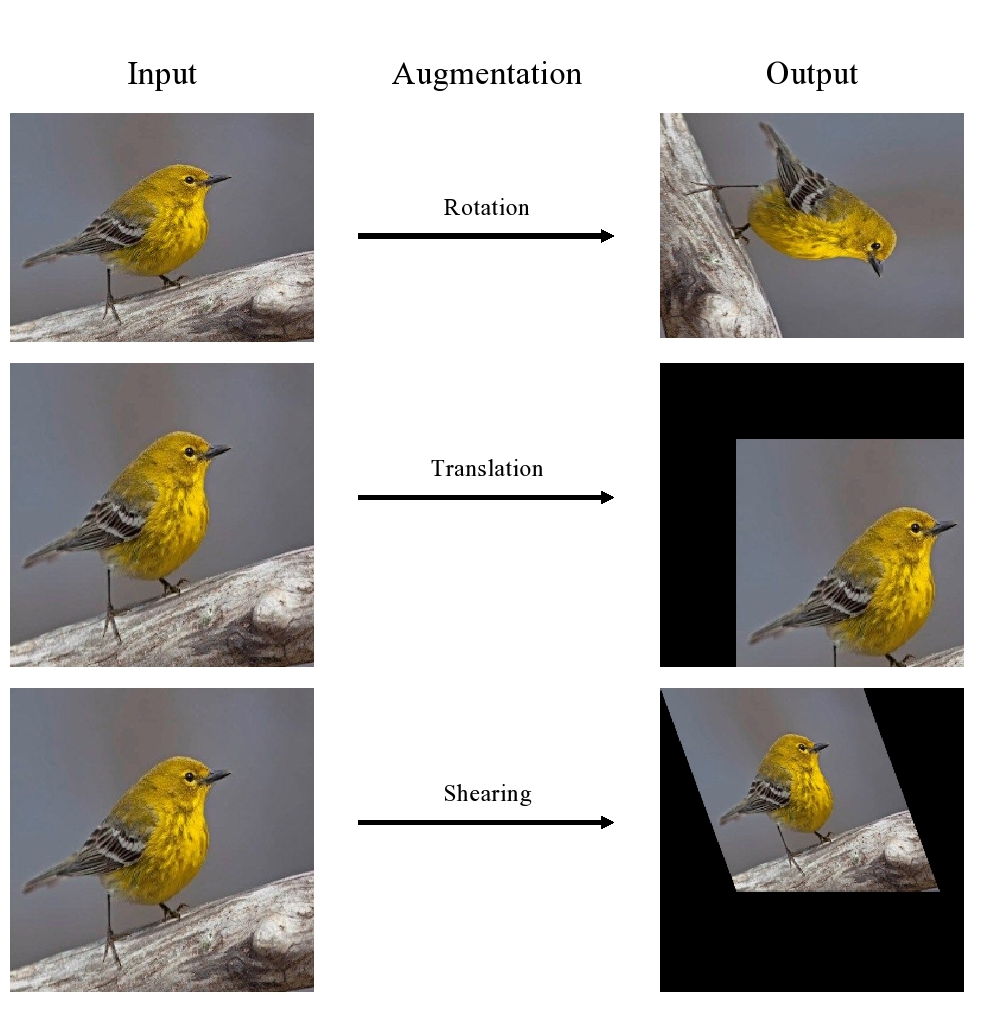}}
\caption{Overview of the geometric data augmentations.}
\label{fig:geometric}
\end{figure}

\item {\textbf{Translation:}}
Translation data augmentation involves shifting an image in any of the upward, downward, right, or left directions, as illustrated in Figure~\ref{fig:geometric}, in order to provide a more diverse representation of the data. The magnitude of this type of augmentation must be selected with caution, as an excessive shift can result in a substantial change in the appearance of the image. For example, translating a digit 8 to the left by half the width of the image could result in an augmented image that resembles the digit 3. Hence, it is imperative to consider the nature of the dataset when determining the magnitude of the translation augmentation to ensure its efficacy.

\item {\textbf{Shearing:}} Shearing data augmentation involves shifting one part of an image in one direction, while the other part is shifted in the opposite direction. This technique can provide a new and diverse perspective on the data, thereby improving the robustness of a model. However, excessive shearing can cause significant deformation of the image, making it difficult for the model to accurately recognize the objects within it. It is therefore important to consider the amount of shearing applied to the data carefully in order to avoid over-augmenting the images and introducing unwanted noise. In this way, shearing can be a powerful tool for enhancing the generalization ability of computer vision models, while avoiding the potential drawbacks of over-augmentation. For example, applying excessive shearing on cat image during data augmentation may result in a distorted, stretched appearance, hindering the ability of a model to correctly classify the image as a cat. It is crucial to find a balance between the amount of shearing applied and the desired level of diversity, as excessive shearing can introduce significant noise.
\end{enumerate}
\textbf{Non-Geometric Data Augmentations}
The non-geometric data augmentation category focuses on modifications to the visual characteristics of an image, as opposed to its geometric shape. This includes techniques such as noise injection, flipping, cropping, resizing, and color space manipulation, as illustrated in Figure~\ref{fig:non-geometric}.
These techniques can help improve the generalization performance of a model by exposing it to a wider variety of image variations during training. However, it is important to consider the trade-off between augmenting the data and preserving the integrity of the underlying information in the image. The following section outlines several classical non-geometric data augmentation approaches.

\begin{enumerate}[label=(\roman*)]
\item {\textbf{Flipping:}} Flipping is a type of image data augmentation technique that involves flipping an image either horizontally or vertically. The efficacy of this method has been demonstrated on various widely-used datasets, including cifar10 and cifar100~\cite{krizhevsky2009learning}. However, care must be taken when applying this technique, as the outcome may depend on the nature of the dataset. For instance, the horizontal flipping of the digit "2" in the Urdu digits dataset~\cite{khan2022introducing} may result in the appearance of the digit "6". As such, the choice of flipping must be made carefully to ensure that the desired level of augmentation is achieved without introducing significant noise into the data.

\item {\textbf{Cropping and resizing:}} Cropping is a common pre-processing data augmentation technique that can be applied randomly or to the center of the image. This technique involves trimming the image and then resizing it back to its original size, preserving the original label of the image. However, caution must be exercised when using cropping as a data augmentation method, as it may result in misleading information for the model, such as cropping the upper or lower part of the digit "8" and making it appear as the digit "0".

\item {\textbf{Noise Injection:}} Noise injection is a data augmentation technique that has been demonstrated to enhance the robustness of neural networks in learning features and defending against adversarial attacks. As shown in the survey of nine datasets from the UCI repository~\cite{shorten2019survey}, the use of noise injection has resulted in impressive performance improvements.

\item {\textbf{Color Space:}}
The manipulation of individual channel values within an image, also known as photometric augmentation, is a type of data augmentation that can help control the brightness of the image. Image data typically consists of three channels: Red (R), Green (G), and Blue (B) and has dimensions of Height (H) x Width (W) x Channels (C). By altering the values of each channel separately, this technique can prevent a model from becoming biased towards specific lighting conditions.  The most straightforward approach to perform color space augmentation involves replacing a single channel within the image with a randomly generated channel of the same size, or with a channel filled with either 0 or 255.  The utilization of color space manipulation is commonly observed in photo editing applications, where it is used to adjust the brightness or darkness of the image~\cite{shorten2019survey}.

\begin{figure}[htbp]
{\includegraphics[width=0.5\textwidth]{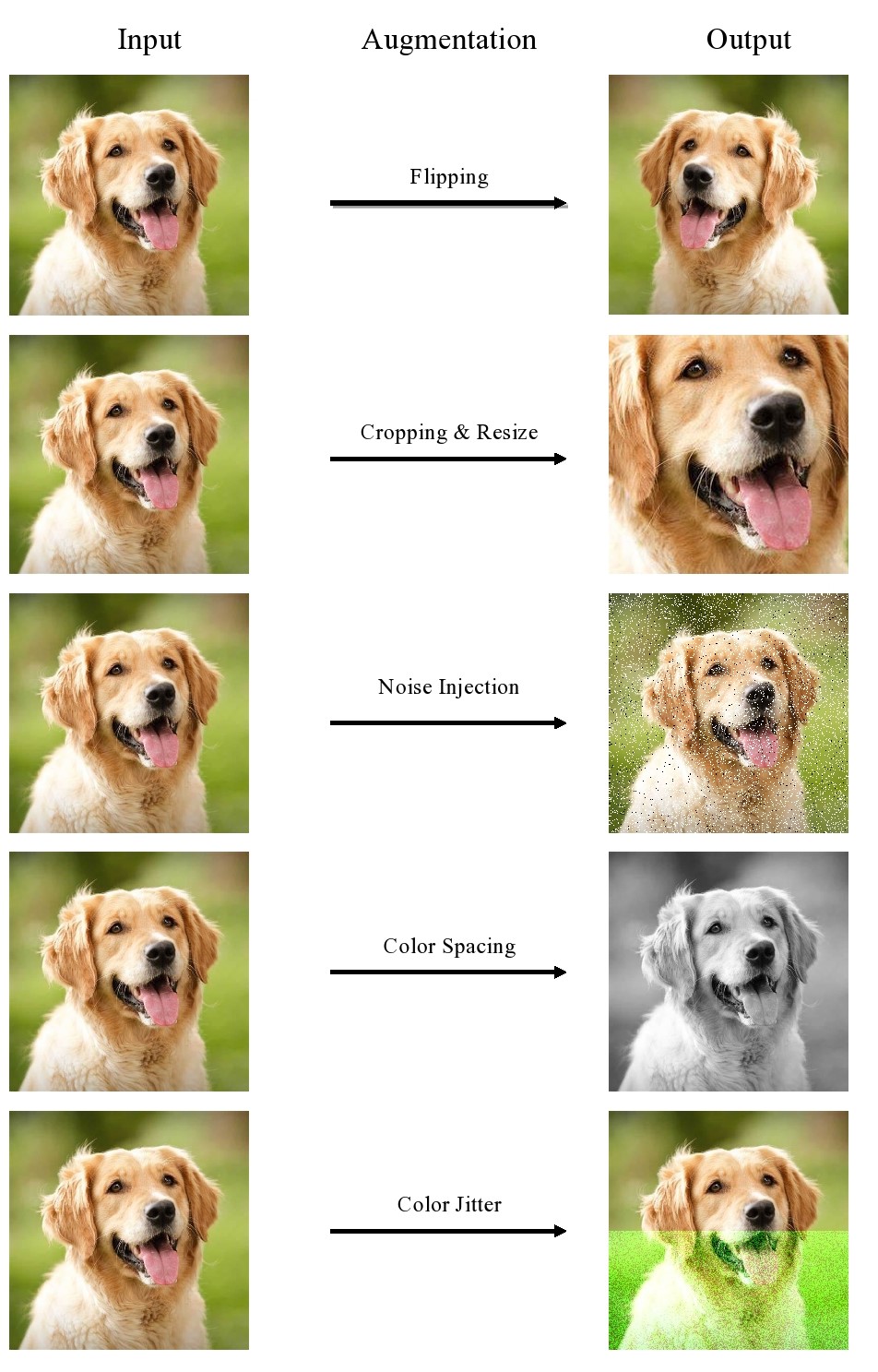}}
\caption{Overview of the non-geometric data augmentations.}
\label{fig:non-geometric}
\end{figure}

\item {\textbf{Jitter:}} Jitter is a technique of data augmentation that involves randomly altering the brightness, contrast, saturation, and hue of an image. The four hyperparameters, i.e., brightness, contrast, saturation, and hue, can be adjusted by specifying their minimum and maximum range. However, it is important to carefully select these ranges as improper adjustments can negatively impact the image's content. For example, increasing the brightness of X-Ray images used for lung disease detection can result in the whitening and blending of the lungs in the X-Ray, hindering the diagnosis of the disease.

\item {\textbf{Kernel Filter: }} Kernel filtering is a form of data augmentation that enhances or softens the image. This is achieved by applying a window, with a specified size of n x n, containing a Gaussian-blur or an edge filter to the image. The Gaussian-blur filter serves to soften the image, while the edge filter sharpens its edges either horizontally or vertically.

\end{enumerate}
\subsubsection{\textbf{Image Erasing Data Augmentations}}
\begin{figure}[htbp]
{\includegraphics[width=0.5\textwidth]{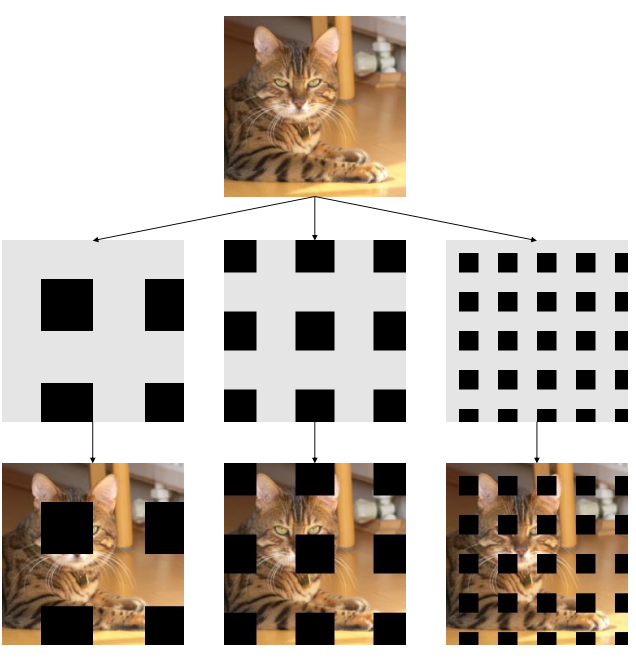}}
\caption{This figure shows the procedure of GridMask augmentation. They produce a mask and then multiply it with the input image. Image is taken from~\cite{chen2020gridmask}.}
\label{fig:gridmask}
\end{figure}

The data augmentation technique of image erasing involves the process of removing specific parts of an image and replacing them with either 0, 255, or the mean of the entire dataset. This type of data augmentation includes various methods such as cutout, random erasing, hide-and-seek, and grid mask, each with their unique implementation and purpose.
\begin{enumerate}[label=(\roman*)]
\item  \textbf{Cutout :} The Cutout data augmentation method involves the random removal of a sub-region within an image, which is then filled with a constant value such as 0 or 255, during the training phase. This approach has been shown to result in improved performance on widely used datasets~\cite{devries2017improved}. An illustration of the Cutout data augmentation process is provided in Figure~\ref{fig:cutmix}.

\item \textbf{Random erasing :} Random Erasing (RE)~\cite{zhong2020random} randomly erases the sub-region in the image similar to cutout. But the main difference is, it randomly determines whether to mask out region or not and also determines the aspect ratio and size of the masked region. RE demonstration for different tasks is shown in figure~\ref{fig:randomerasing}. 

\begin{figure}[htbp]
    \centering
    \includegraphics[width=0.45\textwidth]{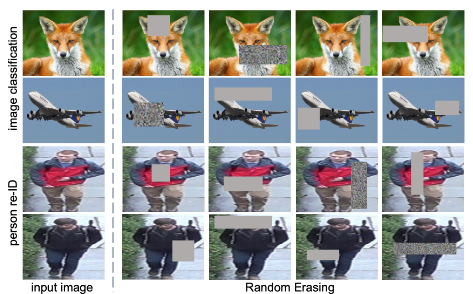}
    \caption{Random erasing examples for different tasks~\cite{zhong2020random}.}\label{fig:randomerasing}
\end{figure}

\item \textbf{Hide-and-Seek}
The process of hide-and-seek data augmentation~\cite{singh2018hide} involves dividing an image into uniform squares of random size and then randomly removing a specified number of these squares. This technique aims to force neural networks to learn relevant features by hiding important information. A different view of the image is presented at each epoch, as depicted in figure~\ref{fig:hide_and_seek}. It is important to note that while this technique has been found to be effective in certain applications, it may also result in the removal of important information which could negatively impact the performance of the model.

\begin{figure}[htbp]
{\includegraphics[width=0.5\textwidth]{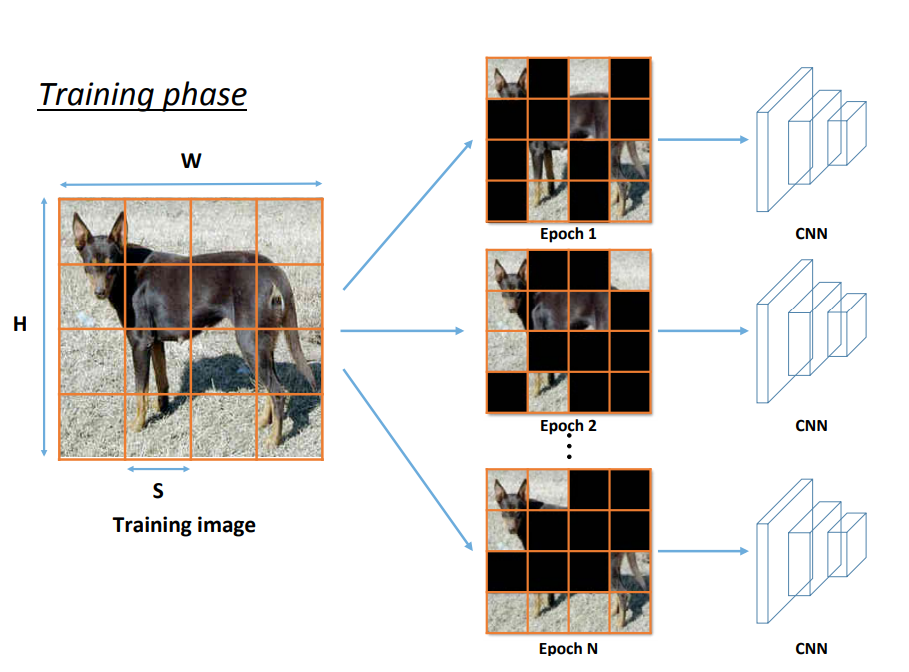}}
\caption{An example of Hide-and-Seek augmentation~\cite{singh2018hide}.}
\label{fig:hide_and_seek}
\end{figure}

\item \textbf{GridMask Data Augmentation}
The GridMask data augmentation technique~\cite{chen2020gridmask} aims to address the challenges associated with randomly removing regions from images. This process, which can completely erase objects or strip away context information, requires a trade-off between the two. To resolve this, GridMask creates a uniform masking pattern and applies it to images as demonstrated in figure ~\ref{fig:gridmask}.
\end{enumerate} 

\subsection{\textbf{Advanced Image Data Augmentations}}
The field of computer vision has seen a surge in interest regarding data augmentation techniques in recent times. This has led to the development of a wide range of innovative methods for augmenting image data, such as mixing images in novel ways, using reinforcement learning, feature-based augmentation, and style-based augmentation. To better understand these advancements, advanced data augmentation techniques have been classified into different major categories. These categories provide a useful framework for surveying the current state of the field and identifying areas for further research and development.
\subsubsection{\textbf{Image Mixing Data Augmentations}}
Image mixing data augmentation has gained popularity in computer vision research in recent years. This technique involves blending one or more images, including the same image, resulting in improved deep neural network model accuracy. We categorize image mixing data augmentation into two sub-categories: single image mixing and non-single image mixing. We compare the effectiveness of these sub-categories on benchmark datasets (such as CIFAR10, CIFAR100, IamgeNet etc), as shown in table ~\ref{classification_results1}, ~\ref{classification_results2}, ~\ref{tab:object_detection_pascal_voc}, ~\ref{tab:object_detection_pascal_voc_2012} and ~\ref{tab:object_detection_results_PASCAL_VOC_1}.

\textbf{ Single Image Mixing Data Augmentations:} A single-image mixing technique uses only one image and mixes a single image from different mixing points of view.  Recently, there has been a lot of work done on single-image augmentation, such as  LocalAugment, SelfAugmentation, SalfMix, and many more. The description of each SOTA single image mixing data augmentation has been discussed below.
\begin{enumerate}[label=(\roman*)]
\item \textbf{Local Augment:} Kim et al.\cite{kim2021local} proposed a technique called LocalAugment, which involves dividing an image into smaller patches and applying different types of data augmentation to each patch. The purpose of this technique is to increase diversity in local features, which could help reduce bias and improve generalization performance of neural networks. While this approach does not preserve the global structure of an image, it provides a rich set of local features that can benefit neural network training. Figure\ref{fig:localaugment} and \ref{fig:localaugment2} provide visual representations of the LocalAugment technique.
Although LocalAugment technique can generate diverse local features of an image, it may not be well-suited for certain types of images that have complex global structures requiring preservation of global spatial relationships. Therefore, it may have some limitations, which should be taken into account while using this technique for image mixing data augmentation.
\begin{figure}[htbp]
{\includegraphics[width=0.5\textwidth]{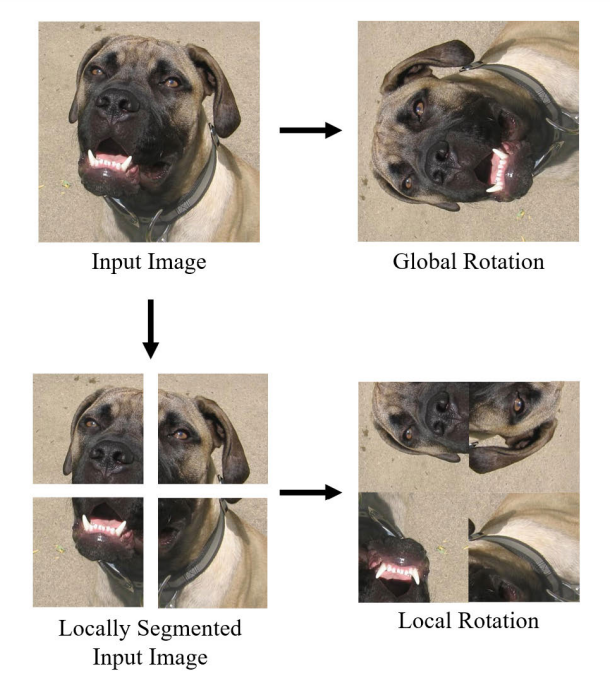}}
\caption{An example of Global and Local Rotation Image, the example is taken from \cite{kim2021local}.}
\label{fig:localaugment}
\end{figure}

\begin{figure}[htbp]
{\includegraphics[width=0.5\textwidth]{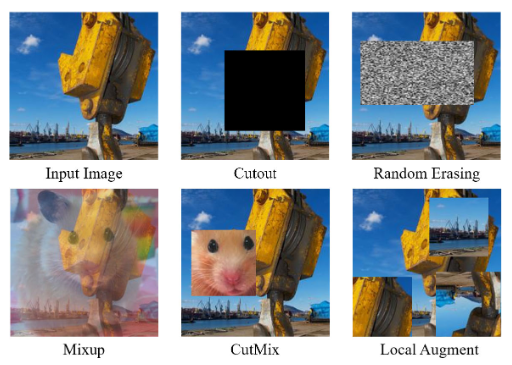}}
\caption{Comparison of LocalAugment with CutOut, MixUp etc, example is taken from~\cite{kim2021local}.}
\label{fig:localaugment2}
\end{figure}

\item \textbf{Self-Augmentation:}This work~\cite{seo2021self} proposes self-augmentation, where a random region of an image is cropped and pasted randomly in the image, improving the generalization capability in few-shot learning. Moreover, the self-augmentation combines regional dropout and knowledge distillation- knowledge from the trained large network is transferred to a small network.The process demonstrated in the figure
 ~\ref{fig:self_augmentation}. 
\begin{figure}[htbp]
{\includegraphics[width=0.5\textwidth]{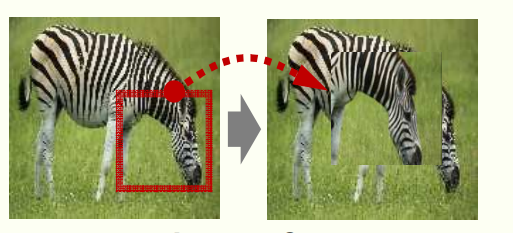}}
\caption{An example of self augmentation, image is taken from\cite{seo2021self}}
\label{fig:self_augmentation}
\end{figure}

\item \textbf{SalfMix:} This work~\cite{choi2021salfmix} focuses on whether it is possible to generalize neural networks based on single-image mixed augmentation. For that purpose,  it proposes SalfMix, the first salient part of the image is found to decide which part should be removed and which portion should be duplicated. Most salient regions are cropped and placed into non-salient regions. This process is defined and compared with other techniques in figure ~\ref{fig:salfmix}.

\begin{figure}[htbp]
{\includegraphics[width=0.5\textwidth]{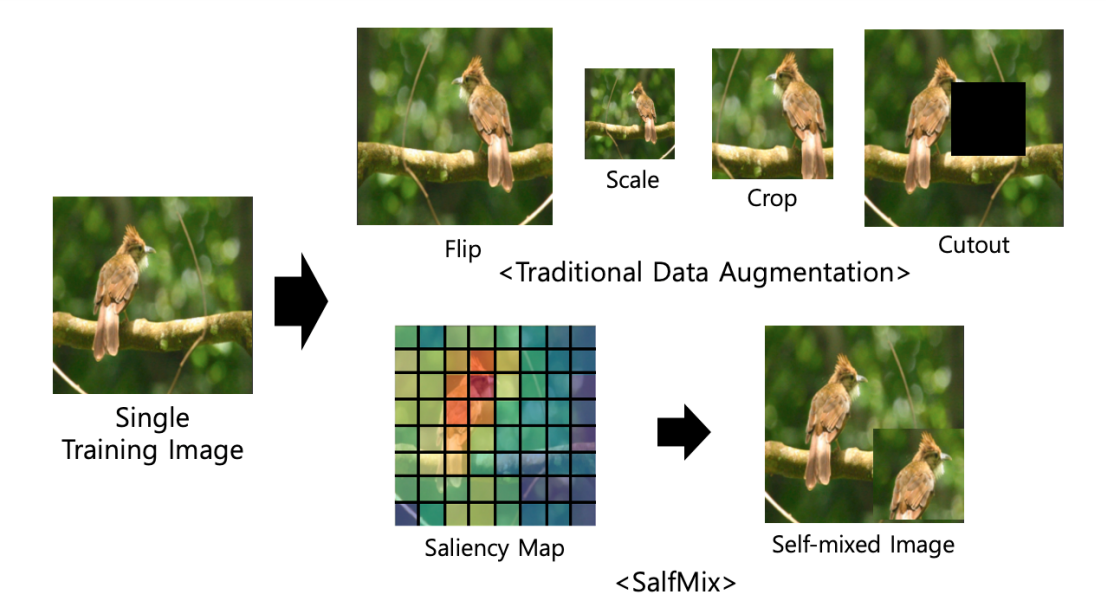}}
\caption{Conceptual comparison between SalfMix method and other single image-based data augmentation methods, the example is taken from~\cite{choi2021salfmix}.}
\label{fig:salfmix}
\end{figure}
\item \textbf{KeepAugment}
KeepAugment~\cite{gong2021keepaugment} is introduced to prevent distribution shift which degrades the performance of neural networks. The idea of KeepAugment is to increase fidelity by preserving the salient features of the image and augmenting the non-salient region. Preserved features help to increase diversity without shifting the distribution. KeepAugment is demonstrated in figure~\ref{fig:keepaugment}.

\begin{figure}[htbp]
{\includegraphics[width=0.5\textwidth]{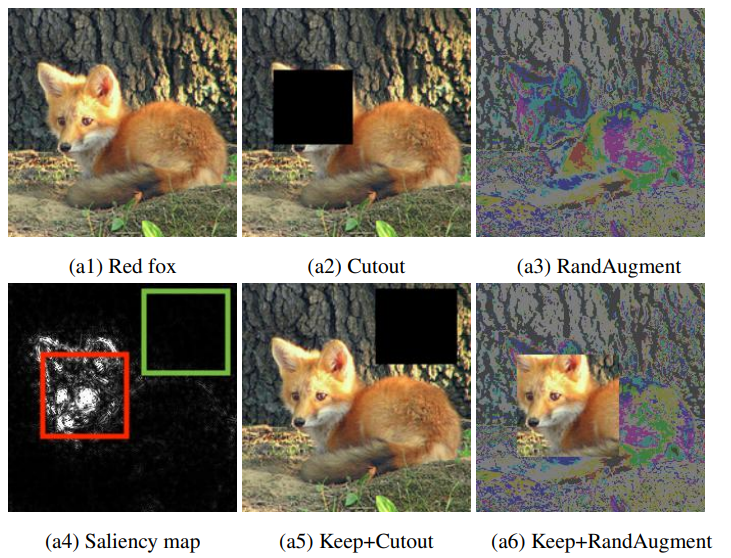}}
\caption{This image shows the example of KeepAugment with other augmentations, courtesy\cite{gong2021keepaugment}. }
\label{fig:keepaugment}
\end{figure}

\item \textbf{You Only Cut Once}
You Only Cut Once (YOCO)~\cite{han2022you} is introduced with the aim of recognizing objects from partial information and improving the diversity of augmentation that encourage neural networks to perform better. YOCO makes two pieces of image and augmentation is applied on each piece, then each piece is concatenated for an image and YOCO shows impressive performance and compared with SOTA augmentations, sometimes it outperforms them. It is easy to implement, has no parameters, and is easy to use. The YOCO augmentation process is shown in figure~\ref{fig:yoco}.

\begin{figure}[htbp!]
{\includegraphics[width=0.5\textwidth]{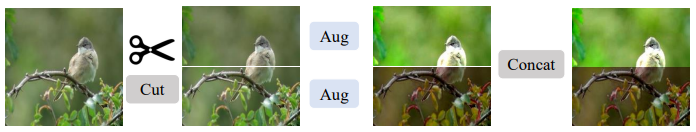}}
\caption{An example of YOCO augmentation, image is taken from~\cite{han2022you}.}
\label{fig:yoco}
\end{figure}

\item \textbf{Cut-Thumbnail:} Cut-Thumbnail~\cite{xie2021cut} is a novel data augmentation, that resizes the image to a certain small size and then randomly replaces the random region of the image with the resized image,  aiming to alleviate the shape bias of the network. The advantage of Cut-thumbnail is, that it not only preserves the original image but also keeps it global in the small resized image. On ImageNet, it shows impressive performance using resnet50.  Overall, the cut-thumbnail process and its comparison are shown in figure~\ref{fig_thumbnails} and figure~\ref{fig:comparison_thumbnail}, respectively.

\begin{figure}[htbp]
{\includegraphics[width=0.5\textwidth]{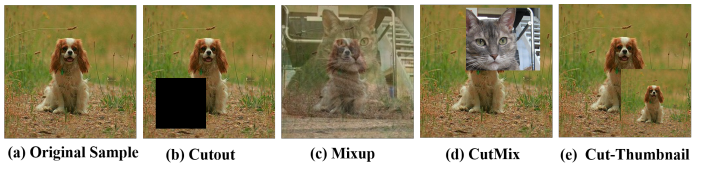}}
\caption{Comparison between existing data augmentation
methods with Cut-Thumbnail, the example is from\cite{xie2021cut}.}
\label{fig:comparison_thumbnail}
\end{figure}
\begin{figure}[htbp]
{\includegraphics[width=0.5\textwidth]{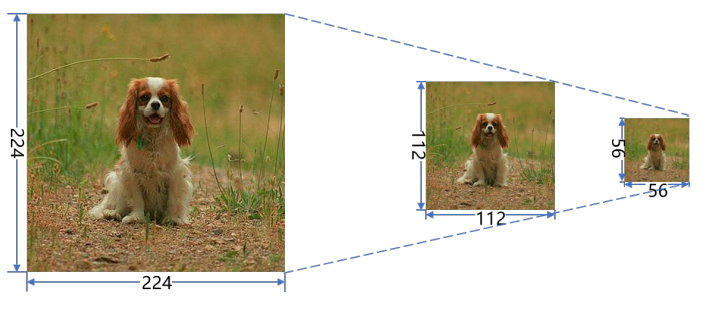}}
\caption{This image shows an example of reduced images that are called thumbnails. After reducing the image to a certain size of 112×112 or 56×56, The dog is still recognizable even though lots of local details are lost, courtesy~\cite{xie2021cut}.}
\label{fig_thumbnails}
\end{figure}
\end{enumerate}
\textbf{ Multi-Images Mixing:}
Multi-Images Mixing data augmentation uses more than one image and applies different mixing strategies. Recently, many researchers have explored a lot of Multi-Images Mixing strategies and still, it is a very attentive topic for many researchers. Recently work has included Mixup, CutMix, SaliencyMix, and many more. Each of the relevant non-single image mixing data augmentation techniques is discussed below.
\begin{enumerate}[label=(\roman*)]
\item \textbf{Mixup: }
Mixup blends two images based on the blending factor  (alpha) and the corresponding labels of these images are also mixed in the same way. Mixup data augmentation~\cite{zhang2017mixup} consistently improved the performance not only in terms of accuracy but also in terms of robustness. Experiments on ImageNet-2012~\cite{russakovsky2015imagenet}, CIFAR-10, CIFAR-100, Google commands~\footnote{https://research.googleblog.com/2017/08/
launching-speech-commands-dataset.html} and UCI datasets~\footnote{http://archive.ics.uci.edu/ml/index.php} showed impressive results on SOTA methods. Further demonstration and comparison are shown in the figure~\ref{fig:cutmix}. 
\item \textbf{CutMix : } CutMix tackles the issues of information loss and region dropout~\cite{yun2019cutmix}. It is inspired by cutout~\cite{devries2017improved}, where any random region is filled with 0 or 255, while in cutmix instead of filling the random region with 0 or 255, the region is filled with a patch from another image. Correspondingly, their labels are also mixed proportionally to the number of pixels mixed. It is compared with other methods and shown in figure~\ref{fig:cutmix}. 

\begin{figure}[htbp]
{\includegraphics[width=0.5\textwidth]{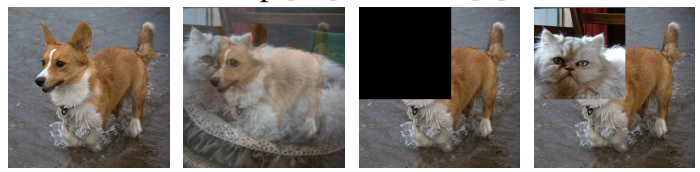}}
\caption{Overview of the Mixup, Cutout, and CutMix, Example is from\cite{yun2019cutmix}.}
\label{fig:cutmix}
\end{figure}

\item {\textbf{SaliencyMix:}} This technique addresses the problem of cutmix and argues that filling a random region of the image with a patch from another will not guarantee that patch has rich information and thereby mixing labels of unguaranteed patches leads the model to learn unnecessary information about the patch~\cite{uddin2020saliencymix}. To deal with that issue, saliencyMix first selects the salient part of the image and pastes it to a random region or salient  or non-salient of another image. It is shown in figure ~\ref{fig:saliencymix1} and figure ~\ref{fig:saliencymix2}.

\begin{figure}[htbp]
{\includegraphics[width=0.5\textwidth]{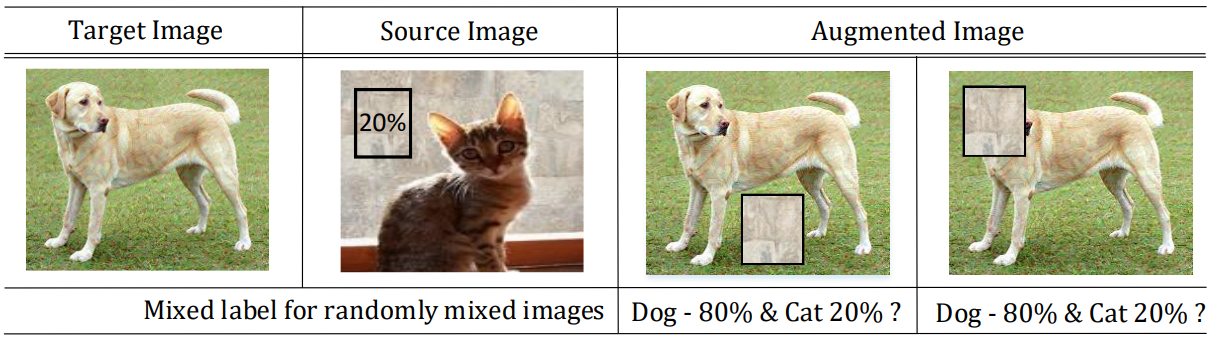}}
\caption{An example of SaliencyMix augmentation, image is taken from~\cite{uddin2020saliencymix}.}
\label{fig:saliencymix1}
\end{figure}
\begin{figure}[htbp]
{\includegraphics[width=0.5\textwidth]{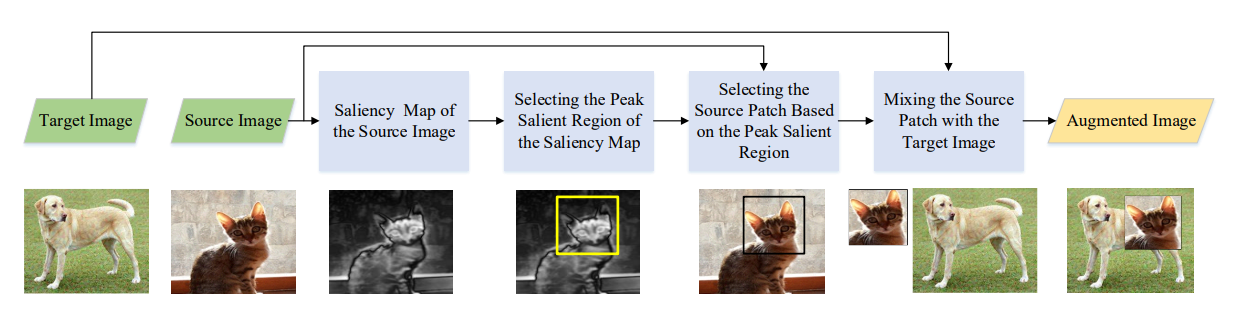}}
\caption{This image shows the proposed SaliencyMix data augmentation procedure, courtesy~\cite{uddin2020saliencymix}}
\label{fig:saliencymix2}
\end{figure}

\item \textbf{RSMDA: Random Slices Mixing Data Augmentation:} RSMDA~\cite{kumar2023rsmda} addresses issues of feature losing in single image erasing data augmentation. RSMDA gets the slices of one image and mixes them with another image alternatively and the corresponding labels are also mixed accordingly. This work further investigates three different strategies of RSMDA; row-wise slice mixing, column-wise slice mixing and randomness of both. Row-wise slice mixing has shown superior performance.  Demonstration of each of the slices mixing strategy is in figure~\ref{fig:rsmda}.   

\begin{figure}[htbp]
    \centering
    {\includegraphics[width=0.5\textwidth]{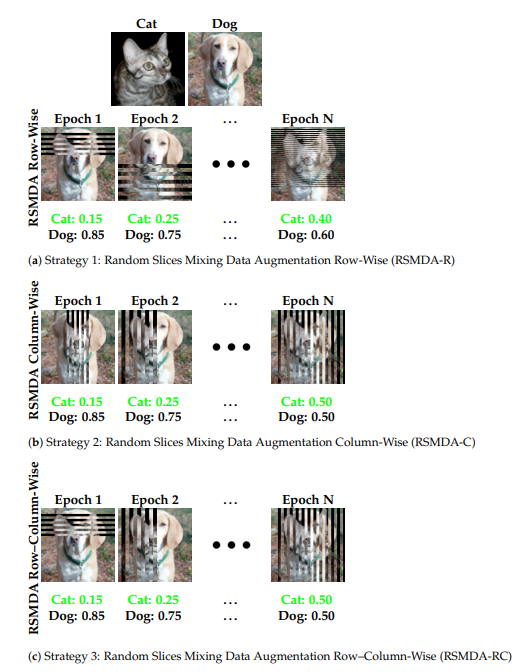}}
    \caption{RSMDA three different strategies. Image is taken from~\cite{kumar2023rsmda}.}
    \label{fig:rsmda}
\end{figure}

\item{\textbf{Puzzle Mix:}} This article~\cite{kim2020puzzle} proposes a puzzle mix data augmentation technique that focuses on using explicitly salient information and basic statistics of image wisely with the aim of breaking the misleading supervision of neural networks over existing data augmentations. Furthermore, the demonstration is shown and compared with relevant methods in figure~\ref{fig:puzzlemix}.

\begin{figure}[htbp]
{\includegraphics[width=0.5\textwidth]{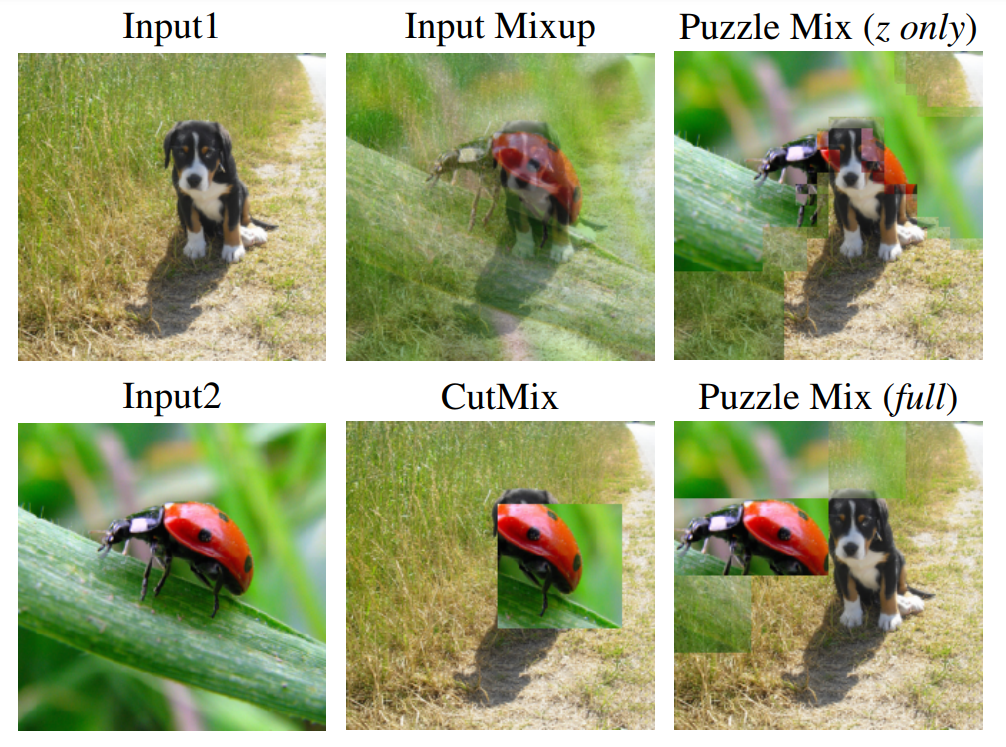}}
\caption{A visual comparison of the mixup methods. Puzzle Mix
ensures to contain sufficient target class information while preserving the local statistics of each in, the example is from~\cite{kim2020puzzle}.}
\label{fig:puzzlemix}
\end{figure}

\item{\textbf{SnapMix:}} The article~\cite{huang2021snapmix} proposes the Semantically Proportional Mixing (SnapMix) that utilises class activation map (CAM) to reduce the label noise level. SnapMix creates the target label considering the actual salient pixel taking part in the augmented image, which ensures semantic correspondence between the augmented image and mixed labels. The overall process is demonstrated and compared with closely matching augmentations in the figure~\ref{fig:snapmix}.
\begin{figure}[htbp]
{\includegraphics[width=0.5\textwidth]{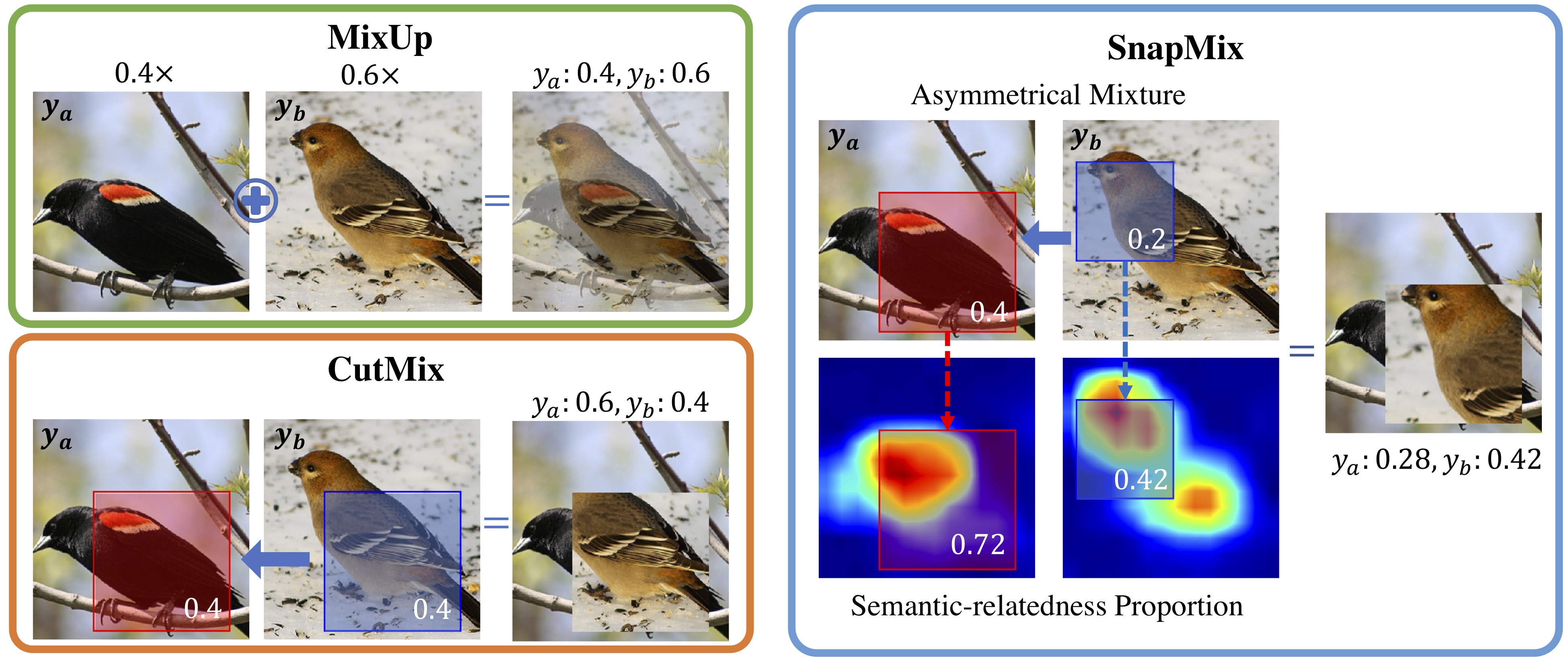}}
\caption{A visual comparison of Mixup, CutMix, and SnapMix. The figure gives an example where a label generated by SnapMix  is visually more consistent with the mixed image semantic structure compared to CutMix and Mixup, courtesy~\cite{huang2021snapmix}.}
\label{fig:snapmix}
\end{figure}

\item {\textbf{FMix:}} This article proposes the FMix~\cite{harris2020fmix}, a kind of mixed sample data augmentation (MSDA), that utilises the random binary masks. These random binary masks are acquired by applying a threshold to low-frequency images that are obtained from the Fourier space. Once the mask is obtained, one color region is applied to input one and another color region is applied to another input.   The overall process is shown in figure~\ref{fig:fmix}. 

\begin{figure}[htbp]
{\includegraphics[width=0.5\textwidth]{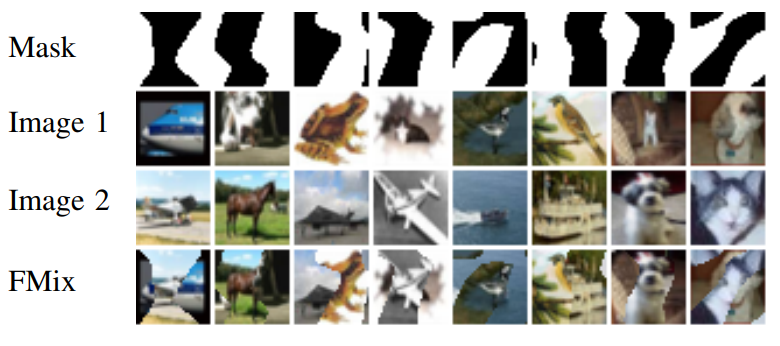}}
\caption{Example masks and mixed images from CIFAR-10 for FMix, example is from~\cite{harris2020fmix}.}
\label{fig:fmix}
\end{figure}

\item {\textbf{MixMo:}}
This paper~\cite{rame2021mixmo} focuses on the learning of multi-input multi-output via sub-network. The main motivation of the paper is to replace direct hidden summing operations with more solid mechanisms. For that purpose, it proposes MixMo, which embeds M inputs into the shared space, mixes and passes these to a further layer for classification. Moreover, the overall process is demonstrated in figure ~\ref{fig:mixmo}:
\begin{figure}[htbp]
{\includegraphics[width=0.5\textwidth]{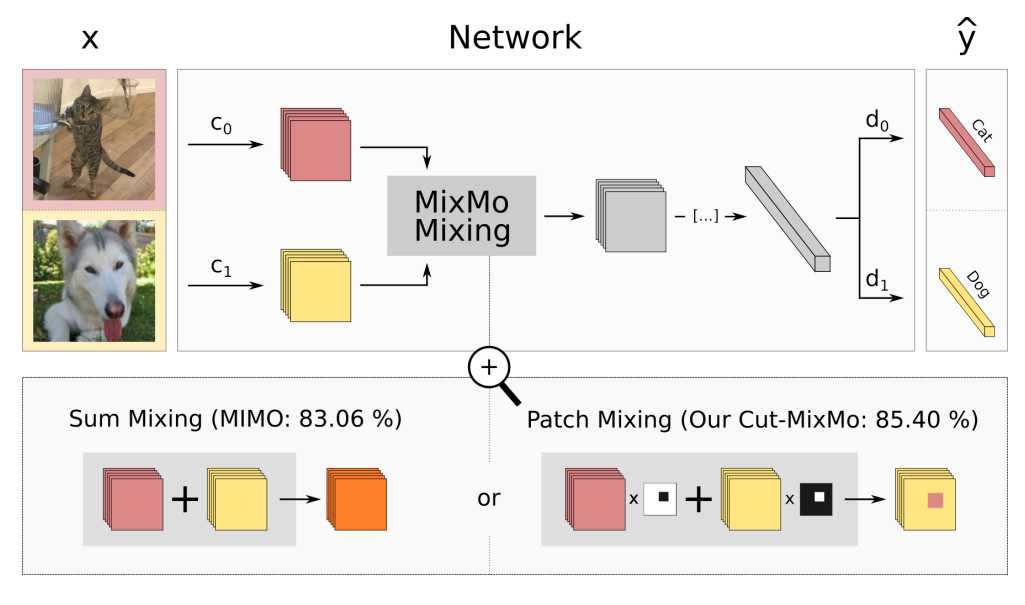}}
\caption{This image shows the overview of MixMo augmentation, the image is taken from\cite{rame2021mixmo}.}
\label{fig:mixmo}
\end{figure}

\item {\textbf{StyleMix:}}
This paper~\cite{hong2021stylemix} targets previous approaches problems that are unable to differentiate between content and style features, approaches such as mixup based data augmentations. To remedy this problem, it  proposes two approaches styleMix and StyleCutMix, this is the first work that separately deals with content and style features of images very carefully and it showed impressive performance on popular benchmark datasets. The overall process is defined and compared with SOTA approaches in figure~\ref{fig:stylemix}.

\begin{figure}[htbp]
{\includegraphics[width=0.5\textwidth]{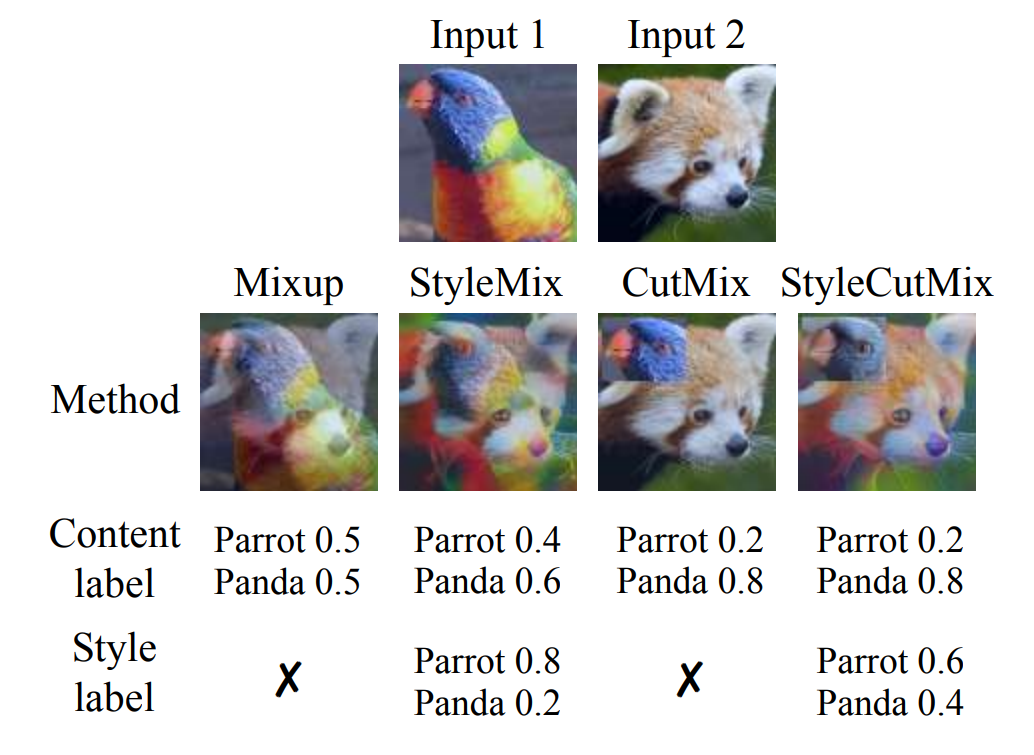}}
\caption{A Visual comparison of StyleMix~\cite{hong2021stylemix} and StyleCutMix with Mixup~\cite{zhang2017mixup} and CutMix~\cite{yun2019cutmix}, example is from~\cite{hong2021stylemix}.}
\label{fig:stylemix}
\end{figure}

\item {\textbf{RandomMix:}} This work~\cite{liu2022randommix} improves generalization capability by proposing randomMix, which randomly selects augmentation from a set of image mixing augmentations and applies it to images, enabling the model to look at diverse samples. This method showed impressive results over SOTA image mixing methods. The overall demonstration is shown in figure~\ref{fig:randommix}.
\begin{figure}[htbp]
{\includegraphics[width=0.5\textwidth]{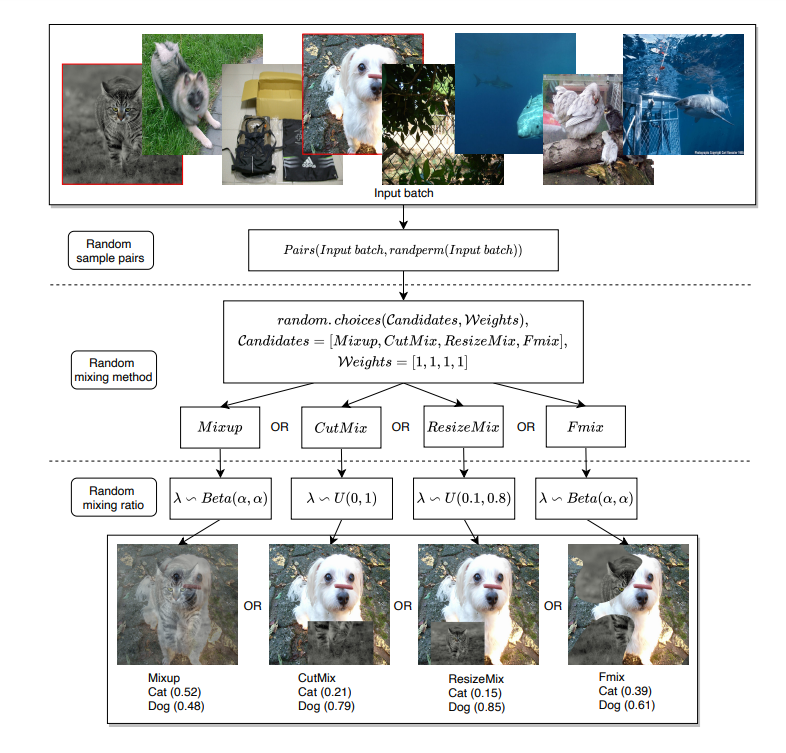}}
\caption{An illustrative example of RandomMix, image is taken from~\cite{liu2022randommix}.}
\label{fig:randommix}
\end{figure}

\item {\textbf{MixMatch:}}
MixMatch data augmentation technique is very useful in semi-supervised learning. MixMatch~\cite{berthelot2019mixmatch} augments single image K times and passes all K number of images to a classifier, averages their prediction and finally, their predictions are sharpened by adjusting their distribution temperature term. It is demonstrated in figure~\ref{fig:mixmatch}.

\begin{figure}[htbp]
{\includegraphics[width=0.5\textwidth]{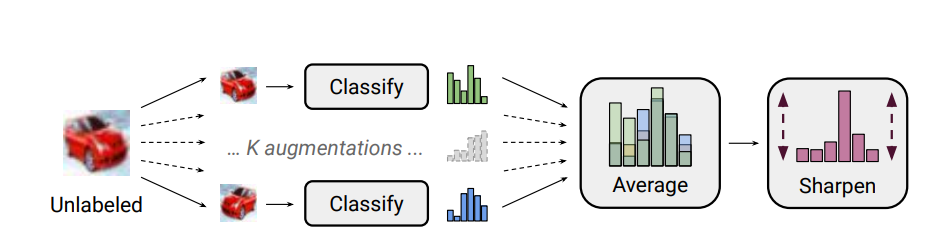}}
\caption{Diagram of the label guessing process used in MixMatch, courtesy~\cite{berthelot2019mixmatch}.}
\label{fig:mixmatch}
\end{figure}

\item {\textbf{ReMixMatch:}}
In this work~\cite{berthelot2019remixmatch}, an extension of MixMatch~\cite{berthelot2019mixmatch} is proposed to make prior work efficient by introducing distribution alignment and augmentation anchoring. The distribution alignment task aims to minimize the gap between the marginal distribution of predictions on unlabeled data and the marginal distribution of ground truth labels. On the other hand, augmentation anchoring feeds multiple strongly augmented versions of the input into the model and encourages each output to be close to the prediction for a weakly-augmented version of the same input. The process is illustrated in figure~\ref{fig:remixmatch}.

\begin{figure}[htbp]
{\includegraphics[width=0.5\textwidth]{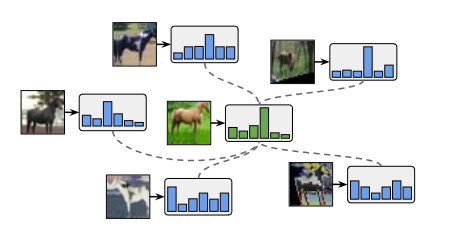}}
\caption{Anchoring augmentation. It makes predictions on strong augmentations of the same image (blue) using the forecast for a weakly enhanced image (green, centre), courtesy~\cite{berthelot2019remixmatch}.}
\label{fig:remixmatch}
\end{figure}

\item {\textbf{FixMatch:}}
Fixmatch~\cite{sohn2020fixmatch} is a method for improving the performance of semi-supervised learning (SSL). It first assigns pseudo-labels to unlabeled images that have a predicted probability above a certain threshold, and then trains the model to match these labels using cross-entropy loss on a strongly augmented version of the image. The process is illustrated in Figure~\ref{fig:fixmatch}.

\begin{figure}[htbp]
{\includegraphics[width=0.5\textwidth]{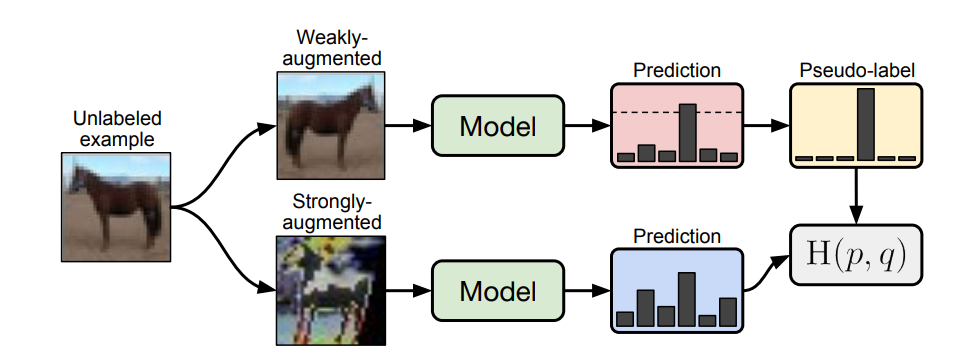}}
\caption{This image shows the procedure of FixMatch, image is taken from\cite{sohn2020fixmatch}.}
\label{fig:fixmatch}
\end{figure}

\item {\textbf{AugMix:}}
The work proposed in~\cite{hendrycks2019augmix} presents Augmix, a data augmentation technique that aims to reduce the distribution gap between training and test data. Augmix applies M random augmentations to an input image, each with a random strength, and merges the resulting images to produce a new image that spans a wider area of the input space. The process is illustrated in Figure~\ref{fig:augmix}, where three branches perform separate augmentations and additional operations are added to increase diversity. The resulting images are then mixed to produce a final augmented image, which is effective in improving model robustness.

\begin{figure}[htbp]
{\includegraphics[width=0.5\textwidth]{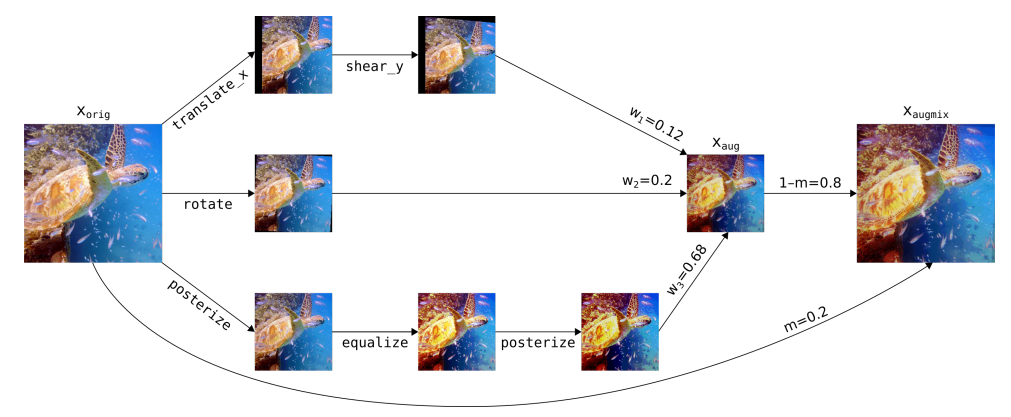}}
\caption{An overall procedure of AugMix augmentation and example is from~\cite{hendrycks2019augmix}.}
\label{fig:augmix}
\end{figure}

\item {\textbf{Simple Copy-Paste is a Strong Data Augmentation Method, for Instance Segmentation:}} The proposed approach in this work~\cite{ghiasi2021simple} involves copying and pasting instances from one image to another to create an augmented image. This simple technique has shown promising results and is easy to implement. Figure~\ref{fig:simple_copy_paste} illustrates the process, where instances from two images are pasted onto each other at different scales.
\begin{figure}[htbp]
{\includegraphics[width=0.5\textwidth]{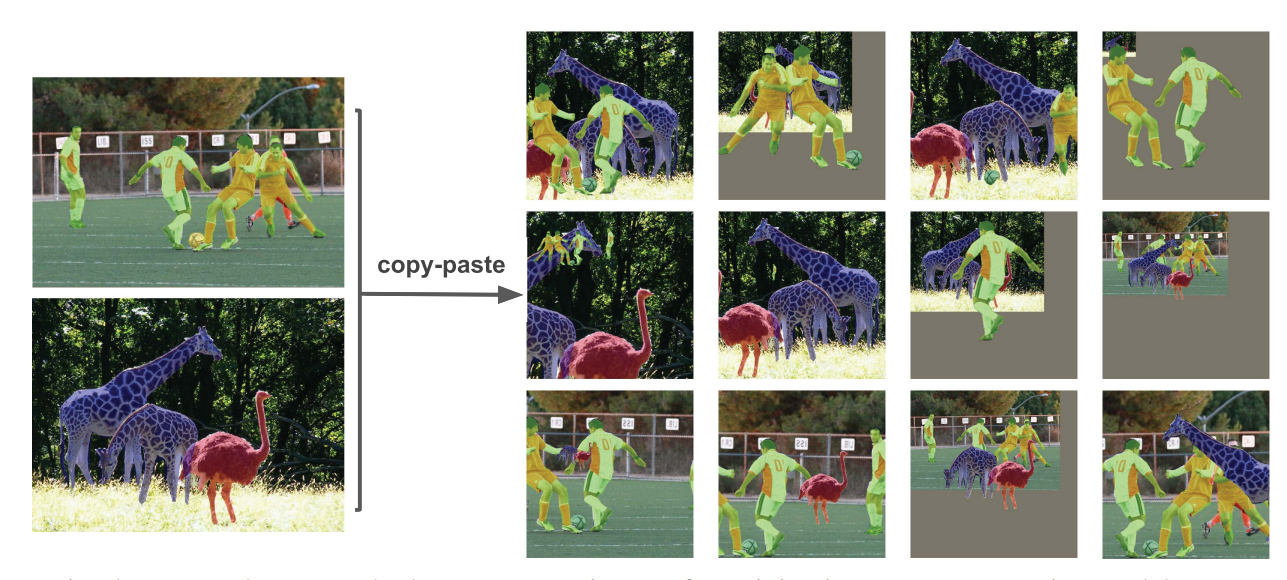}}
\caption{Image augmentation performed by simple Copy-Paste method, image courtesy~\cite{ghiasi2021simple}.}
\label{fig:simple_copy_paste}
\end{figure}

\item {\textbf{Improved Mixed-Example Data Augmentation:}} Recently, label non-preserving data augmentation techniques based on linear combinations of two examples have demonstrated promising results. In this paper~\cite{summers2019improved}, the authors investigate two research questions: (i) the reasons behind the success of these methods and (ii) the significance of linearity in data augmentations. Figure~\ref{fig:improved_mixed_example} illustrates the overall process.
\begin{figure}[htbp]
{\includegraphics[width=0.5\textwidth]{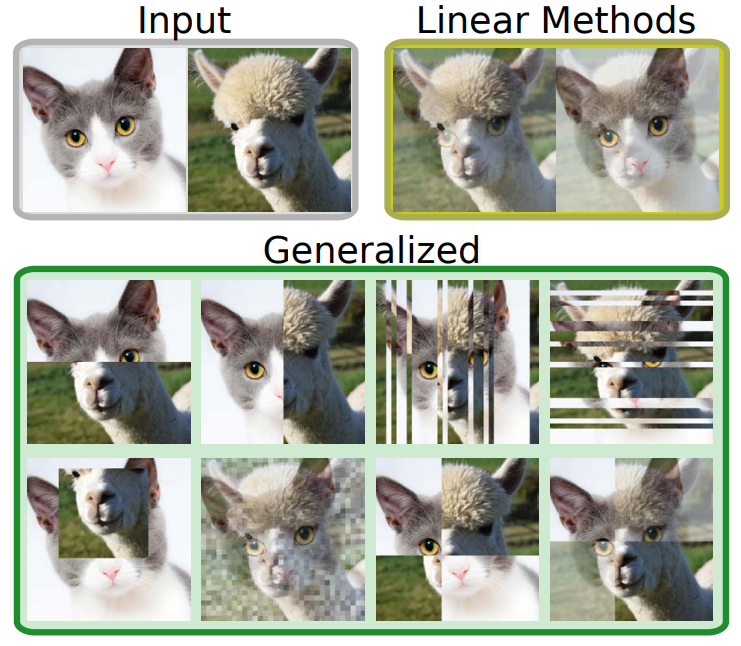}}
\caption{A visual comparison of linear methods and generalized augmentation performed by Improved Mixed-Example, image is taken from~\cite{summers2019improved}.}
\label{fig:improved_mixed_example}
\end{figure}

\item {\textbf{Random image cropping and patching (RICAP):}} Random Image Cropping and Patching (RICAP)~\cite{takahashi2018ricap} is a new data augmentation technique that cuts and mixes four images rather than two images. The key idea behind RICAP is to crop patch from each of the four images and then mixes these patch to create augmented image. The labels of the images are also mixed in proportion to the area of the patcehs. This technique showed impressive performance on popular datasets i.e. CIFAR10, CIFAR100, and imageNet. RICAP demonstration  is shown in figure ~\ref{fig:ricap}.

\begin{figure}[htbp]
{\includegraphics[width=0.5\textwidth]{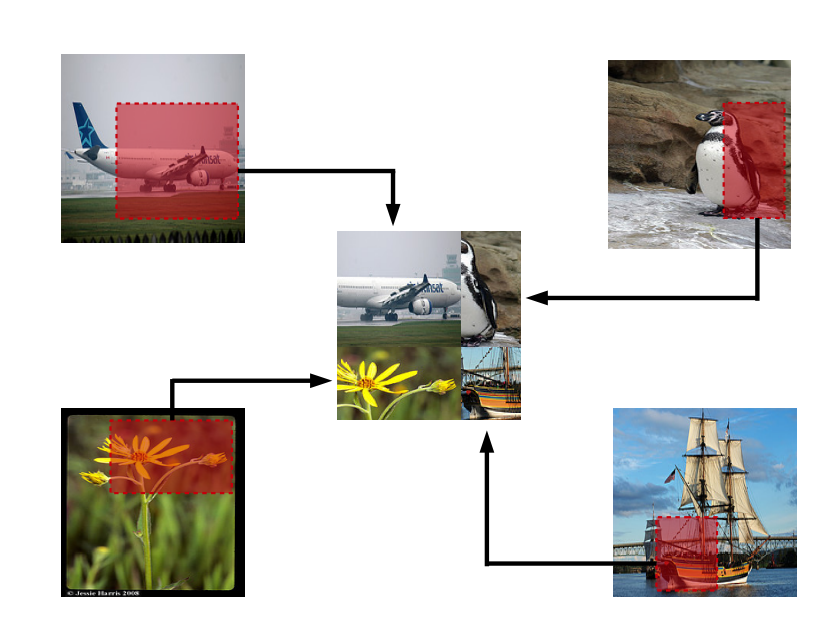}}
\caption{A conceptual explanation of the RICAP data augmentation, the example is from~\cite{takahashi2018ricap}.}
\label{fig:ricap}
\end{figure}

\item {\textbf{Cutblur:}} This article~\cite{yoo2020rethinking} explores and analyses existing data augmentation techniques for super-resolution and proposes another data augmentation technique for super-resolution, named cutblur that cuts high-resolution image patches and pastes to corresponding low-resolution images and vice-versa. Cutblur shows impressive performance on several super-resolution benchmark datasets. Furthermore, the process is illustrated in figure ~\ref{fig:rethinking1} and ~\ref{fig:rethinking2}.

\begin{figure}[htbp]
{\includegraphics[width=0.5\textwidth]{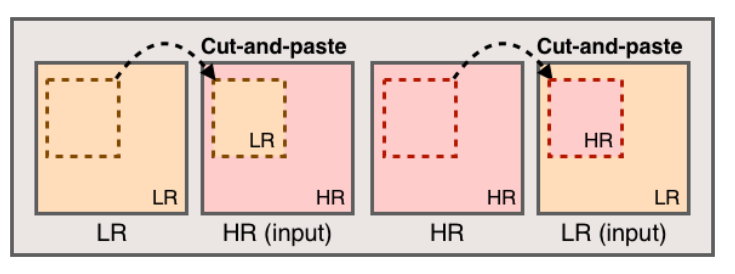}}
\caption{An Schematic illustration of CutBlur operation, image is taken from~\cite{yoo2020rethinking}.}
\label{fig:rethinking1}
\end{figure}

\begin{figure}[htbp]
{\includegraphics[width=0.5\textwidth]{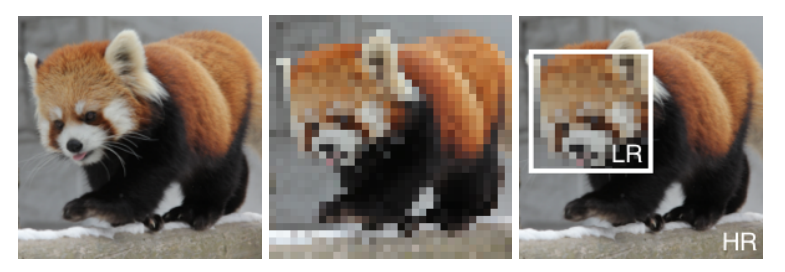}}
\caption{A visual comparison between High resolution, low resolution and CutBlur, courtesy\cite{yoo2020rethinking}.}
\label{fig:rethinking2}
\end{figure}

\item {\textbf{ResizeMix: Mixing Data with Preserved Object Information and True Labels : }}
The ResizeMix~\cite{qin2020resizemix} method directly cuts and pastes the source data in four different ways to target the image. These four different ways include salient part, non-part, random part or resized source image to patch, as shown in the figure~\ref{fig:resizemix}. It addresses two questions:
\begin{itemize}
  \item How to obtain a patch from the source image?
  \item where to paste the patch from the source image in the target image?
\end{itemize}
Furthermore, it was found that saliency information is not important to promote mixing data augmentation. ResizeMix is shown in the figure~\ref{fig:resizemix}.

\begin{figure}[htbp]
{\includegraphics[width=0.5\textwidth]{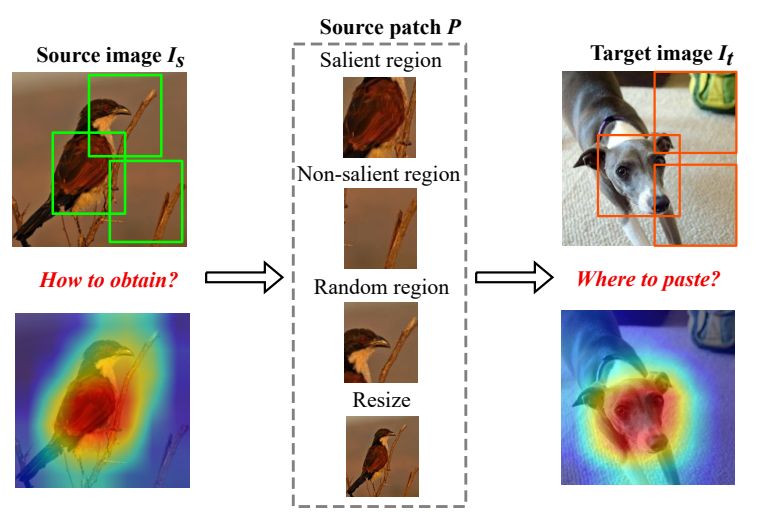}}
\caption{A visual representation of different cropping manners from the source image and different pasting manners to the target image, image is taken from\cite{qin2020resizemix}.}
\label{fig:resizemix}
\end{figure}

\item {\textbf{ClassMix: Segmentation-Based Data Augmentation for Semi-Supervised Learning :}}
This research work~\cite{olsson2021classmix} proposes novel data augmentation for semi-supervised learning for semantic segmentation task. It showed that traditional data augmentations are not effective for image semantic segmentation as they are for image classification. The proposed data augmentation named ClassMix, augments the training sample by mixing unlabeled samples, by exploiting network prediction while taking into account object boundaries. It showed a massive performance gain on two common semantic segmentation datasets for semi-supervised learning. The overall process is shown in the figure~\ref{fig:classmix}.

\begin{figure}[htbp]
{\includegraphics[width=0.5\textwidth]{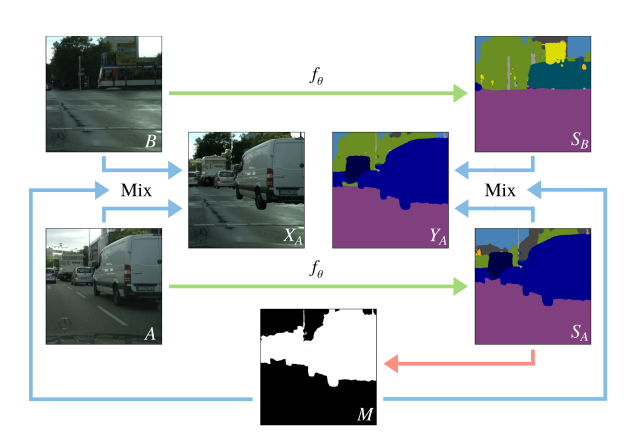}}
\caption{In a visual representation classMix augmentation, two images are sampled then based on the predictions of each image a binary mask is created. The mask is then used to mix the images and their predictions, he image is taken from~\cite{askslsson2021classmix}.}
\label{fig:classmix}
\end{figure}

\item {\textbf{Context Decoupling Augmentation for Weakly Supervised Semantic Segmentation (WSSS):}}
This article~\cite{su2021context} addresses the problem of traditional data augmentation techniques for WSSS, increasing the same contextual data semantic samples does not add much value in object differentiation, i.e. in image classification,  “cat” recognition is due to the cat itself and its surrounding context, these both contexts discourages model to focus only on the cat. To tackle that issue, this work proposes a novel data augmentation, named Context Decoupling Augmentation (CDA).  CDA increases diversity of the specific object and it guides the network to break the dependencies between object and contextual information. In this way, it also provides augmentation and the network focuses on object(s) only rather than object(s) and its contextual information. A comparison of traditional data augmentation and CDA is shown below in the figure~\ref{fig:CDAWSS}.

\begin{figure}[htbp]
{\includegraphics[width=0.5\textwidth]{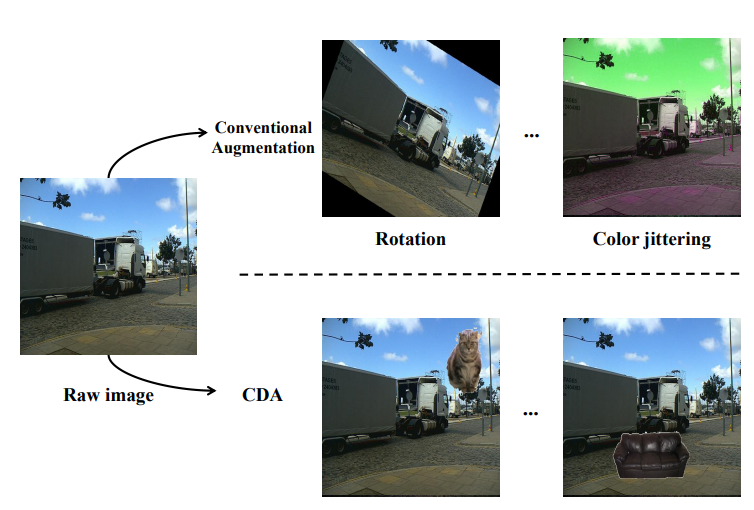}}
\caption{A visual representation of the difference between the conventional augmentation approach and context decoupling augmentation (CDA), image is taken from~\cite{su2021context}.}
\label{fig:CDAWSS}
\end{figure}

\item {\textbf{ObjectAug: Object-level Data Augmentation for Semantic Image Segmentation:}}
This article~\cite{zhang2021objectaug} addresses the problem of mixing image-level data augmentation strategies, which failed to work for segmentation as object and background are coupled and boundaries of objects are not augmented due to their fixed semantic bond with the background. To mitigate this problem, this article~\cite{zhang2021objectaug} proposes a novel approach named ObjectAug, object-level augmentation for semantic segmentation.  First, it separates object(s) and backgrounds from an image with the help of semantic labels then each object is augmented using popular data augmentation techniques such as flipping and rotating.  Pixel changes due to these data augmentations are restored using image inpainting. Finally, the object(s) and background are coupled to create an augmented image. Experimental results suggest that ObjectAug has shown performance improvement for segmentation tasks. Furthermore, ObjectAug is shown in the figure~\ref{fig:ObjectAug}.
\end{enumerate} 
\begin{figure}[htbp]
{\includegraphics[width=0.5\textwidth]{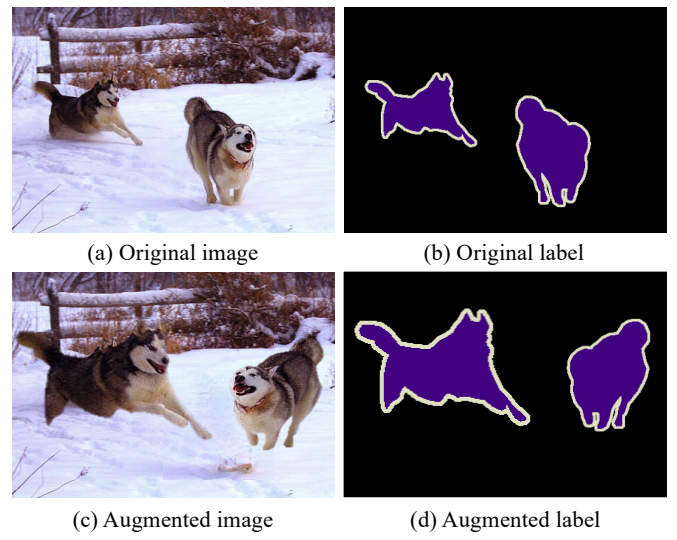}}
\caption{ObjectAug can perform various augmentation methods for each object to boost the performance of semantic segmentation. The left husky is scaled and shifted, while the right one is flipped and shifted. Thus, the boundaries between objects are extensively augmented to boost their performance, the example is from~\cite{zhang2021objectaug}.}
\label{fig:ObjectAug}
\end{figure}

\subsubsection{\textbf{AutoAugment}} The goal of this technique is to find the data augmentation policies from training data. It solves the problem of finding the best augmentation policy as a discrete search problem. It consists of a search algorithm and a search space. Furthermore, these techniques are classified into two sub-categories based on reinforcement learning and non-reinforcement learning. 

\begin{itemize}
    \item Reinforcement learning data augmentation
    \item Non-Reinforcement learning data augmentation
\end{itemize}

\textbf{Reinforcement Learning data augmentations:}
Reinforcement learning data augmentation techniques generalize and improve the performance of deep networks in an environment.
\begin{enumerate}[label=(\roman*)]
\item \textbf{AutoAugment:} This work~\cite{cubuk2019autoaugment} 
automatically finds the best data augmentation rather than manual data augmentation. To address the limitations of manual search-based data augmentation, this article proposes autoaugment, where search space is designed and has policies consisting of many sub-policies. Each sub-policy has two parameters one is the image processing function and the second one is the probability with magnitude.  These sub-policies are found using reinforcement learning as a searching algorithm. The overall process is demonstrated in figure~\ref{fig:rda}.

\begin{figure}[htbp]
{\includegraphics[width=0.5\textwidth]{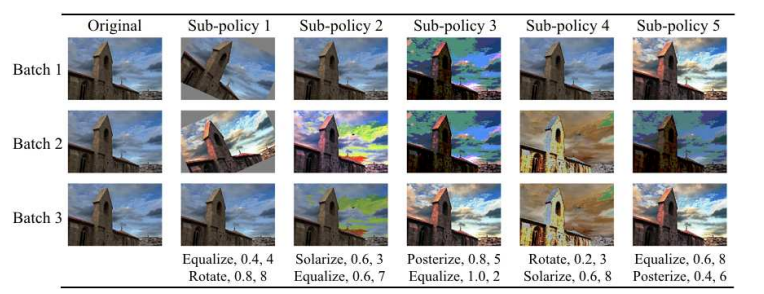}}
\caption{A visual overview of the sub-policies from ImageNet using AutoAugment, example is from\cite{cubuk2019autoaugment}.}
\label{fig:rda}
\end{figure}
\item {\textbf{Fast AutoAugment:}}
Fast Autoaugment~\cite{lim2019fast} addresses the problem of autoaugment, autoaugment takes a lot of time to find the optimal data augmentation strategy. To reduce the searching time, fast auto augment finds more optimal data augmentations using an efficient search strategy based on density matching.  It reduces the higher order of training time compared to autoaugment. The overall procedure is shown in figure~\ref{fig:fast_autoaugment}.

\begin{figure}[htbp]
{\includegraphics[width=0.5\textwidth]{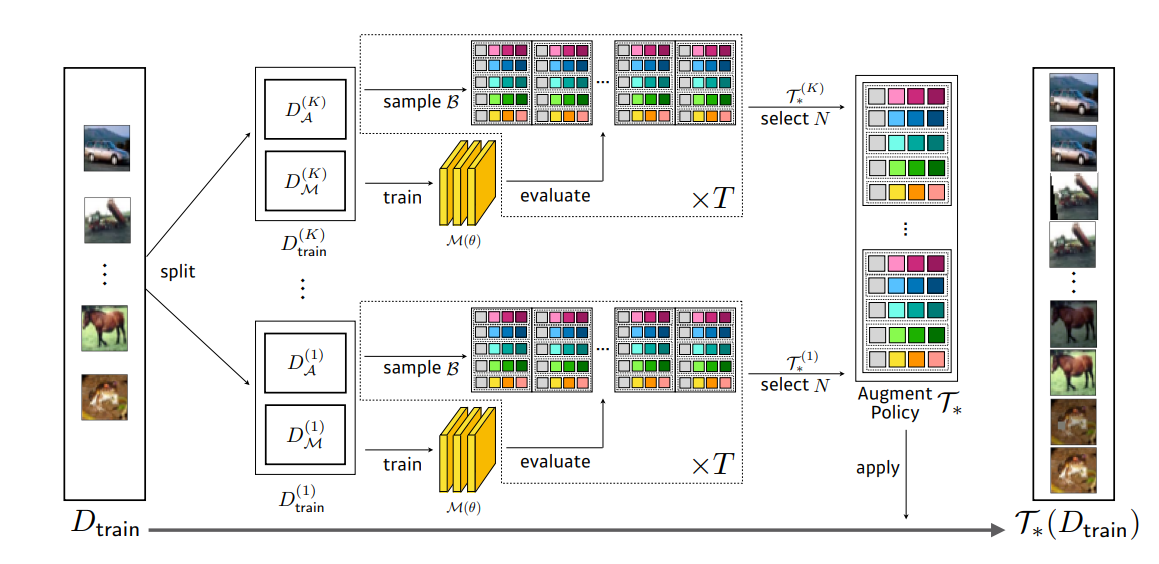}}
\caption{An overall procedure of augmentation search by Fast AutoAugment algorithm, courtesy\cite{lim2019fast}.}
\label{fig:fast_autoaugment}
\end{figure}

\item {\textbf{Faster AutoAugment:}}
This article proposes a faster autoaugment\cite{hataya2020faster} policy intending to find effective data augmentation policies very efficiently. Faster autoaugment is based on a differentiable augmentation searching policy and additionally, it not only estimates gradients for many transformation operations having discrete parameters but also provides a mechanism for choosing operations efficiently. Moreover, it introduces a training objective function with aim of minimising the distance between original and augmented distribution, which is also differentiable. Parameters of augmentations are updated during backpropagation. The Overall process is defined in figure~\ref{fig:faster_autoaugment}:

\begin{figure}[htbp]
{\includegraphics[width=0.5\textwidth]{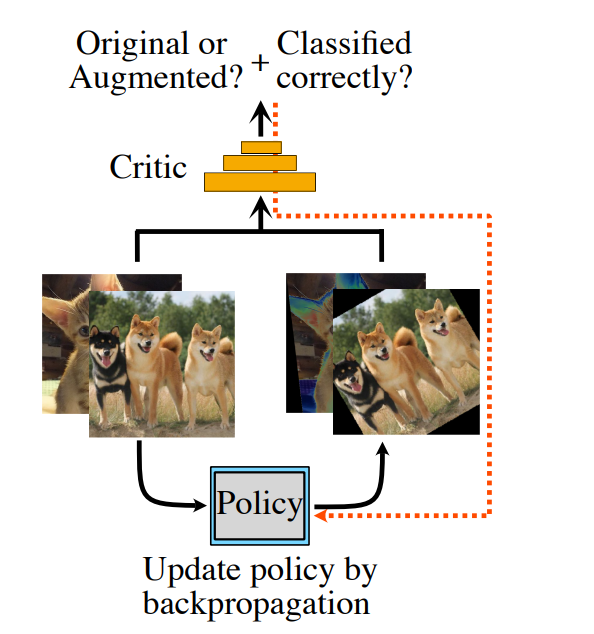}}
\caption{An Overview of the Faster AutoAugment augmentation, image is taken from\cite{hataya2020faster}.}
\label{fig:faster_autoaugment}
\end{figure}

\item {\textbf{Reinforcement Learning with Augmented Data:
}}
This paper proposes Reinforcement Learning with Augmented Data (RAD)~\cite{laskin2020reinforcement}, easily pluggable and enhances the performance of RL algorithms by targeting two issues i) learning data efficiency and ii) generalisation capability for new environments. Furthermore, it shows traditional data augmentation techniques enable RL algorithms to outperform complex SOTA tasks for pixel-based control and state-based control. Overall process is demonstrated in figure~\ref{fig:rad}:

\begin{figure}[htbp]
{\includegraphics[width=0.5\textwidth]{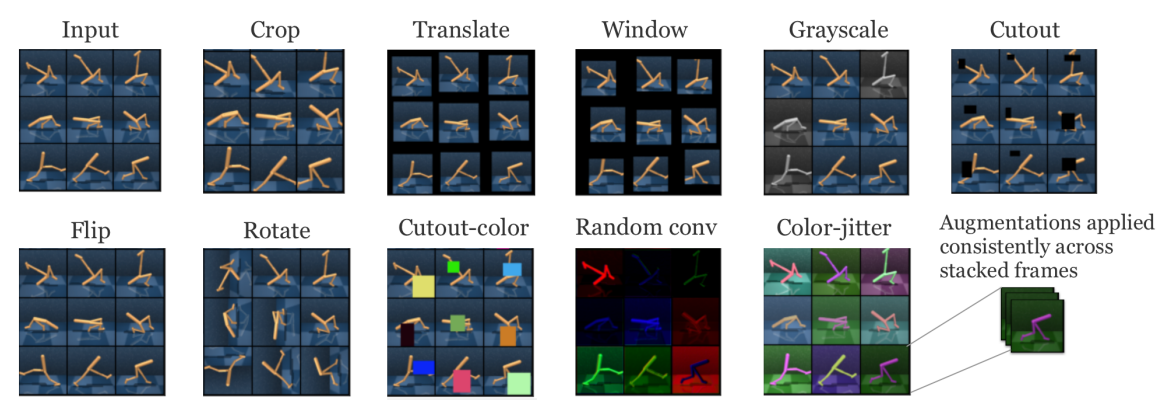}}
\caption{An overview of different augmentation investigated in RAD, the example is from\cite{laskin2020reinforcement}.}
\label{fig:rad}
\end{figure}

\item {\textbf{Local Patch AutoAugment with Multi-Agent Collaboration: }}
This is the first work~\cite{lin2021local} that finds data augmentation policy for patch level using reinforcement learning, named multi-agent reinforcement learning (MARL). MARL starts by dividing images into patches and jointly finds the optimal data augmentation policy for each patch.  It shows competitive results on SOTA benchmarks. Overall process is defined in figure~\ref{fig:localpatch}:
\begin{figure}[htbp]
{\includegraphics[width=0.5\textwidth]{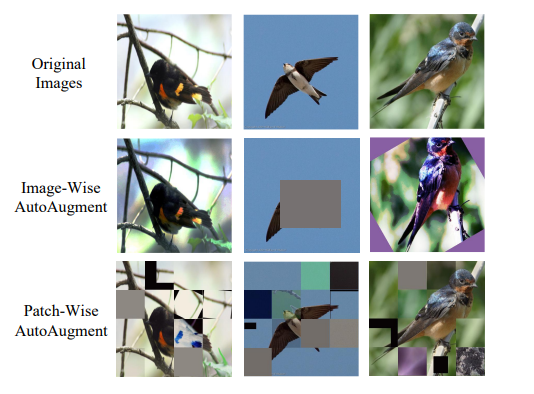}}
\caption{An illustration of different automated augmentation policies, courtesy~\cite{lin2021local}.}
\label{fig:localpatch}
\end{figure}

\item {\textbf{Learning Data Augmentation Strategies for
Object Detection:}}
This work~\cite{zoph2020learning} proposes to use autoaugment that learns the best policies for object detection. It finds the best operation and optimal value. Moreover, it addresses two key issues of augmentation for object detection,
\begin{enumerate}[]  
\item Classification learned policies can not directly be applied for detection tasks, and it adds more complexity to deal with bounding boxes if geometric augmentations are applied.  
\item Most researchers think it adds much less value compared to designing new network architecture so gets less attention but augmentation for object detection should be selected carefully.  
\end{enumerate} 
Some sub-policies for this data augmentation are shown below:

\begin{figure}[htbp]
{\includegraphics[width=0.5\textwidth]{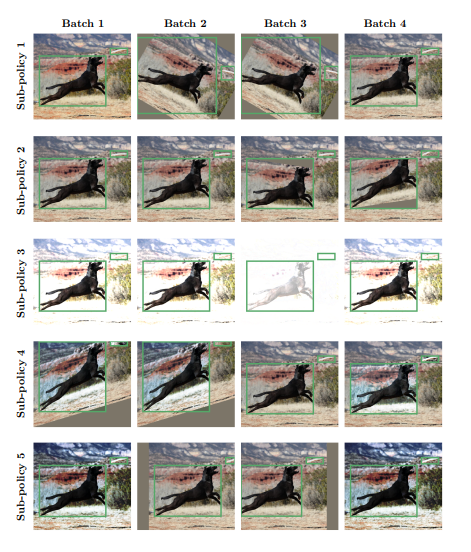}}
\caption{Different data augmentation sub-policies explored, image is taken from~\cite{zoph2020learning}. }
\label{fig:LDAS}
\end{figure}

\text{Sub-policy 1. (Color, 0.2, 8), (Rotate, 0.8, 10)} \\
\text{Sub-policy 2. (BBox\_Only\_ShearY, 0.8, 5)} \\
\text{Sub-policy 3. (SolarizeAdd, 0.6, 8), (Brightness, 0.8, 10)} \\
\text{Sub-policy 4. (ShearY, 0.6, 10), (BBox\_Only\_Equalize, 0.6,8)} \\
\text{Sub-policy 5. (Equalize, 0.6, 10), (TranslateX, 0.2, 2)} \\

\item {\textbf{Scale-aware Automatic Augmentation for Object Detection:}}
This work~\cite{chen2021scale} proposes a new data augmentation for object detection named scale aware autoAug, first, it defines a search space where image level and box level data augmentation are prepared for scale invariance, secondly, it also proposes a new search metric named Pareto scale balance for search augmentation effectively and efficiently. Some examples of data augmentation are shown in figure ~\ref{fig:scale_aware}.

\begin{figure}[htbp]
{\includegraphics[width=0.5\textwidth]{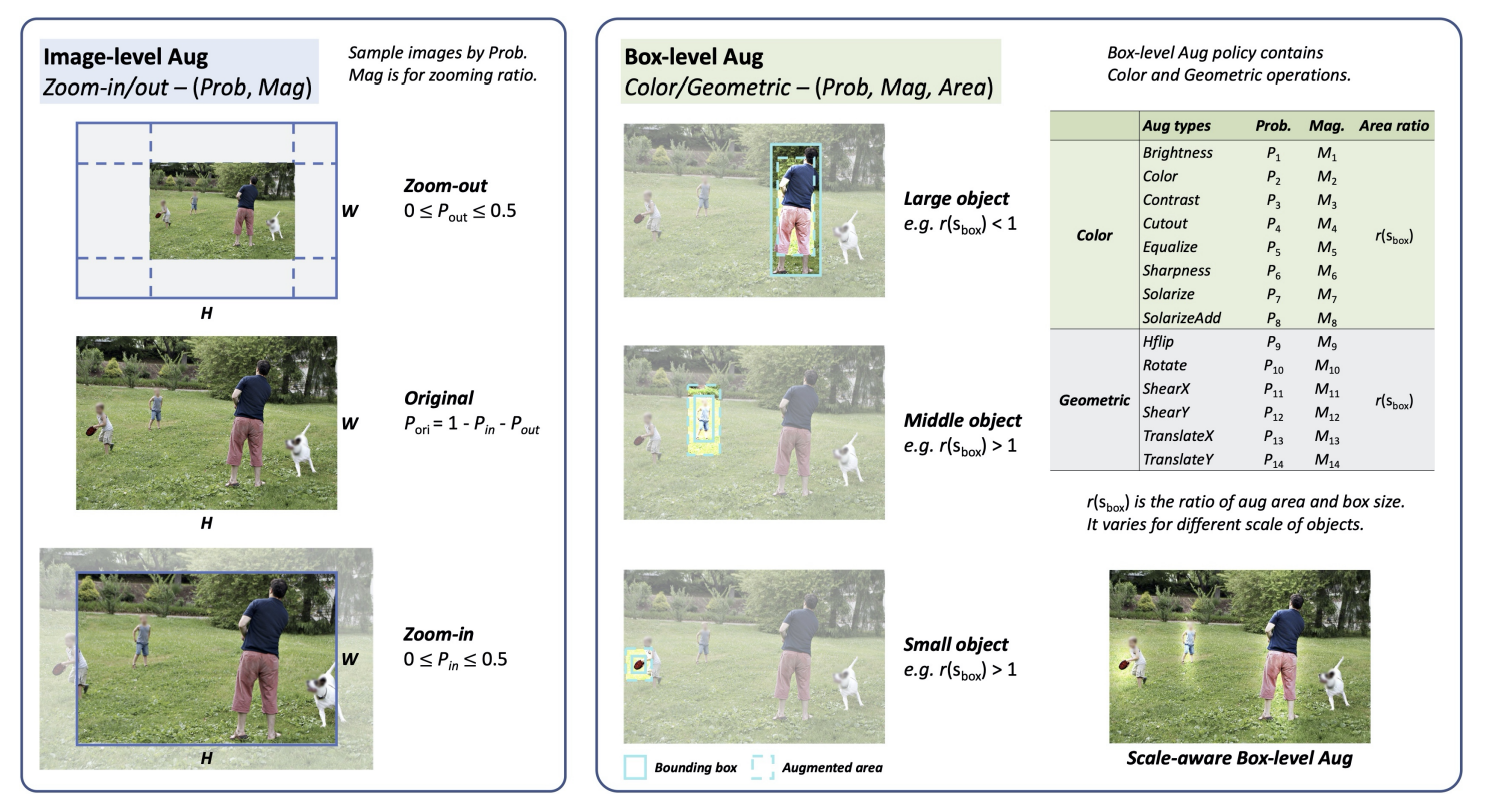}}
\caption{Example of scale-aware search space which includes image level and box-level augmentation, the example is from, \cite{chen2021scale}.}
\label{fig:scale_aware}
\end{figure}

\end{enumerate}
\subsubsection{ \textbf {Non-Reinforcement Learning data augmentations}}
In auto-agument category, there are some approaches that do not require any reinforcement learning algorithm to find the best data augmentation, we refer to them as non-reinforcement learning data augmentation. We categorise a few of them as discussed below. 
\begin{enumerate}[label=(\roman*)]
\item {\textbf{RandAugment: }}
Previous optimal augmentation finding uses reinforcement or some complex learning strategy that takes a lot of time to find. RandAugment augmentation~\cite{cubuk2020randaugment} removes obstacles of a separate searching phase, which makes training more complex and consequently adds computational cost overhead. To break this, randaugment applies randomly N number of data augmentations with M magnitude of all augmentations. Some visualisation is demonstrated in the figure~\ref{fig:randaugment}:
\begin{figure}[htbp]
{\includegraphics[width=0.5\textwidth]{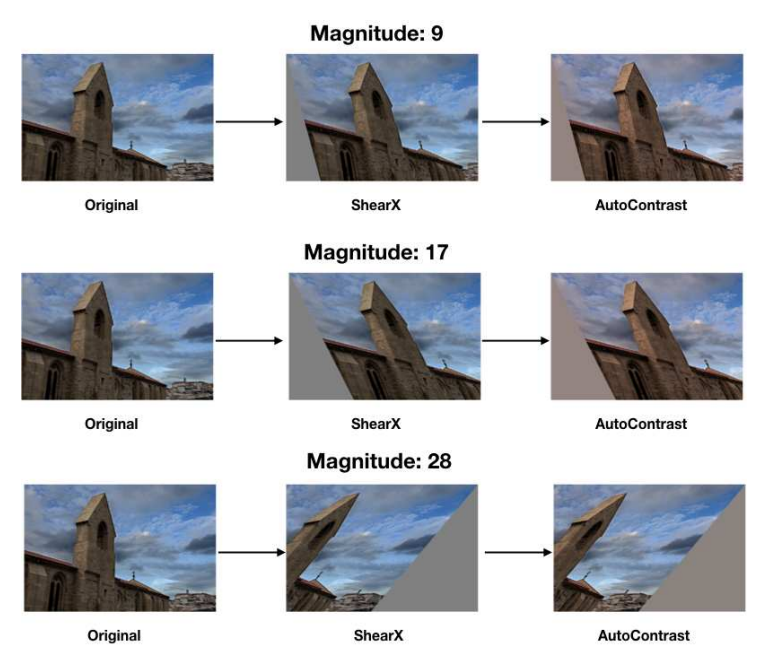}}
\caption{Example images augmented by RandAugment, image is taken from \cite{cubuk2020randaugment}.}
\label{fig:randaugment}
\end{figure}
\item {\textbf{RangeAugment}} RangeAugment~\cite{mehta2022rangeaugment} is a data augmentation technique that aims to improve upon the shortcomings of existing approaches like AutoAugment and RandAugment. These methods use manually-defined ranges of magnitudes for each type of data augmentation, which can result in sub-optimal policies. In contrast, RangeAugment learns efficient ranges of magnitudes for each augmentation and composite data augmentation by introducing an auxiliary loss based on image similarity. This loss is designed to control the magnitude ranges, resulting in more effective and optimal policies. The process of RangeAugment is illustrated in Figure~\ref{fig:rangeaugment}.

\begin{figure}[htbp]
{\includegraphics[width=0.5\textwidth]{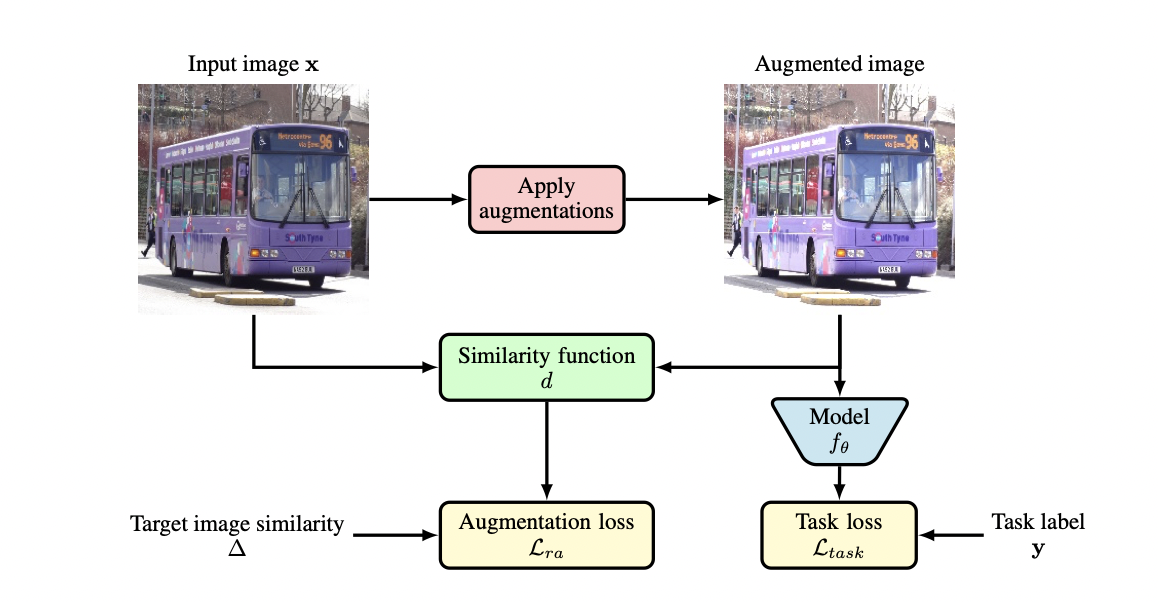}}
\caption{RangeAugment with neural network training \cite{mehta2022rangeaugment}.}
\label{fig:rangeaugment}
\end{figure}

\item {\textbf{ADA: Adversarial Data Augmentation for Object Detection:}}
Data augmentation for object detection has improved performance but it is difficult to understand whether these augmentations are optimal or not. This article~\cite{behpour2019ada} provides a systematic way to find optimal adversarial perturbation of data augmentation from an object detection perspective, that is based on game-theoretic interpretation aka Nash equilibrium of data. Nash equilibrium provides the optimal bounding box predictor and optimal design for data augmentation. Optimal adversarial perturbation refers to the worst perturbation of ground truth, that forces the box predictor to learn from the most difficult distribution of samples. An example is shown in figure~\ref{fig:ADA}.

\begin{figure}[htbp]
{\includegraphics[width=0.5\textwidth]{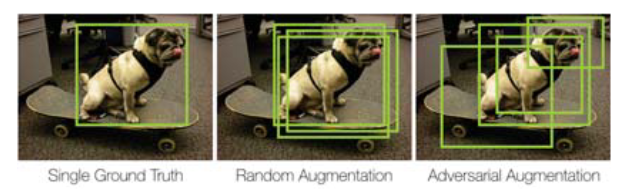}}
\caption{Annotation distribution types. Adversarial augmentation chooses bounding boxes that are as distinct from the truth as possible while yet containing crucial object characteristics. The example is taken from~\cite{behpour2019ada}.}
\label{fig:ADA}
\end{figure}

\item {\textbf{Deep CNN Ensemble with Data Augmentation for Object Detection:}}
This article~\cite{guo2015deep} proposes a new variant of the regions with convolutional neural network (R-CNN) model with two core modifications in training and evaluation. First, it uses several different CNN models as ensembler in R-CNN, secondly, it smartly augments PASCAL VOC training examples with Microsoft COCO data by selecting a subset from Microsoft COCO datasets that are consistent with PASCAL VOC. Consequently, it increases the dataset size and improves the performance. The schematic diagram is shown in the figure~\ref{fig:deep_cnn_objct}.

\begin{figure}[htbp]
{\includegraphics[width=0.5\textwidth]{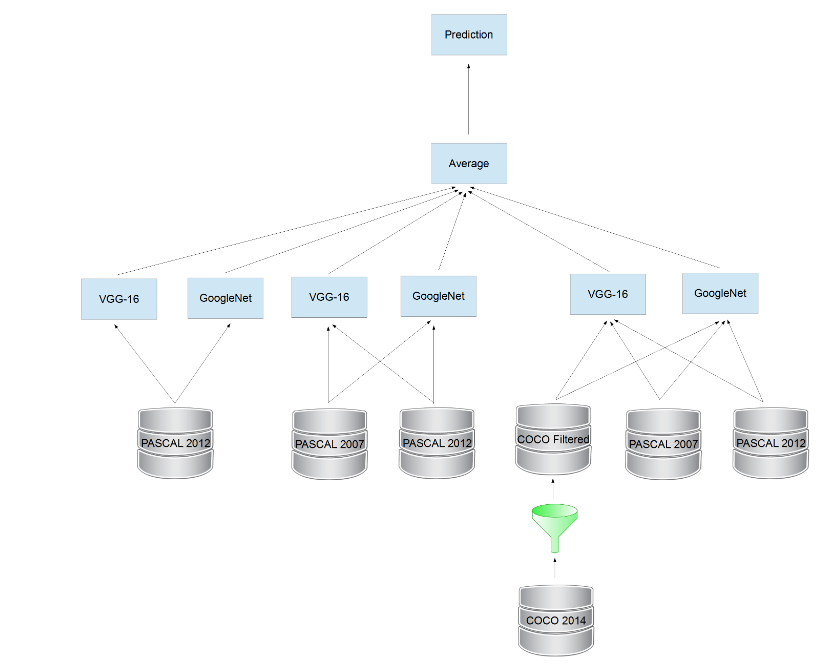}}
\caption{ The proposed schematic diagram. The example is taken from~\cite{guo2015deep}.}
\label{fig:deep_cnn_objct}
\end{figure}

\item {\textbf{Robust and Accurate Object Detection via Adversarial Learning:}}
This article~\cite{chen2021robust} first shows classifier performance gain from different data augmentations when it is fine-tuned to object detection tasks and suggests that the performance in terms of accuracy or robustness is not improving. The article provides a unique way of exploring adversarial samples that helps to improve performance. To do so, it augments the example during the fine-tuning stage for object detectors by exploring adversarial samples, which is considered as model-dependent data augmentation. First, it picks the stronger adversarial sample from detector classification and localization layers and ensures the augmentation policy remains consistent. It showed significant performance gain in terms of accuracy and robustness on different object detection tasks. Furthermore, the robustness and accuracy of the proposed method are shown in figure~\ref{fig:raodl}.

\begin{figure}[htbp]
{\includegraphics[width=0.5\textwidth]{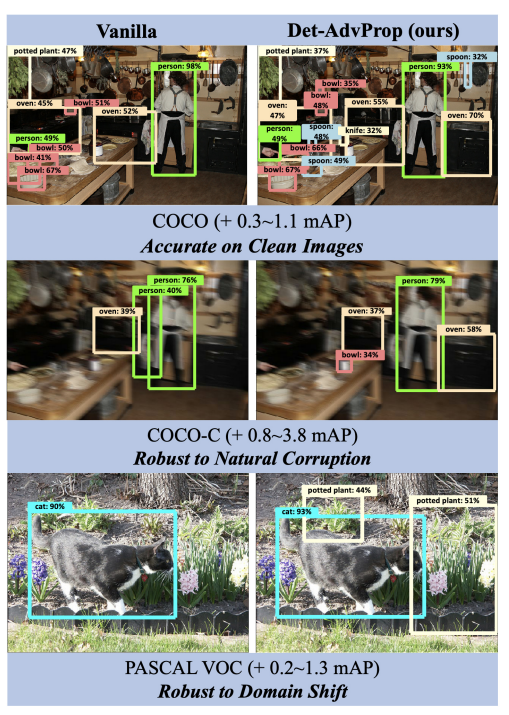}}
\caption{Overview of Robust and Accurate Object detection via adversarial learning. In the top image, it improves object detector accuracy on clean images. In middle, improves the detector's robustness against natural corruption, and at the bottom, it improves the robustness against cross-dataset domain shift. The image is taken from~\cite{chen2021scale}.}
\label{fig:raodl}
\end{figure}

\item {\textbf{Perspective Transformation Data Augmentation for Object Detection:}}
This article~\cite{wang2019perspective} proposes a new data augmentation for objection detection named perspective transformation that generates new images captured at different angles. Thus, it mimics images as if they are taken at a certain angle where the camera can not capture those images. This method showed effectiveness on several object detection datasets. An example of the proposed data augmentation is shown in figure~\ref{fig:PTDA}.

\begin{figure}
    \centering
    {\includegraphics[width=0.5\textwidth]{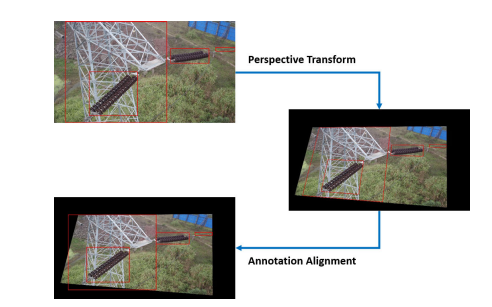}}
    \caption{Perspective transformation data augmentation. An example image is taken from~\cite{wang2019perspective} }
    \label{fig:PTDA}
\end{figure}

\item {\textbf{Deep Adversarial Data Augmentation for Extremely Low Data Regimes:}}
This article~\cite{zhang2020deep} addresses the issue of extremely low data regimes-labeled data is very less, no unlabeled data at all. To deal with that problem, it proposes a deep adversarial data augmentation (DADA), where data augmentation is formulated as a problem of training class conditional and supervised GAN. Furthermore, it also introduces new discriminator loss with aim of fitting data augmentation where real and augmented samples are forced to participate equally and be consistent in finding decision boundaries.

\end{enumerate}

\subsubsection{\textbf{Feature augmentation}}
Feature augmentation is another category of data augmentation, where images are transformed into embedding or representation then data augmentation is performed on the embedding of the image. Recently a few works have been done in this area, we selectively highlight the work in a precise way.
\begin{enumerate}[label=(\roman*)]
\item {\textbf{FeatMatch: Feature-Based Augmentation for Semi-Supervised Learning : }}
This work~\cite{kuo2020featmatch} presents a novel approach of data augmentation in features space for SSL inspired by an image-based SSL method that uses a combination of augmentations of the images and consistency regularization. Image-based SSL methods are restricted to only conventional data augmentation. To break this end, the feature-based SSL method produced diverse features from complex data augmentations. One key point is, these advanced data augmentations exploit the information from both intra-class and inter-class representations extracted via clustering.  The proposed method only showed significant performance gain on min-Imagenet such as an absolute 17.44\% gain on miniImageNet, but also showed robustness on samples that are out-of-distribution. Moreover, the difference between image-level and feature-level augmentation and consistency is shown in figure~\ref{fig:featMatch}.
\begin{figure}[htbp]
{\includegraphics[width=0.5\textwidth]{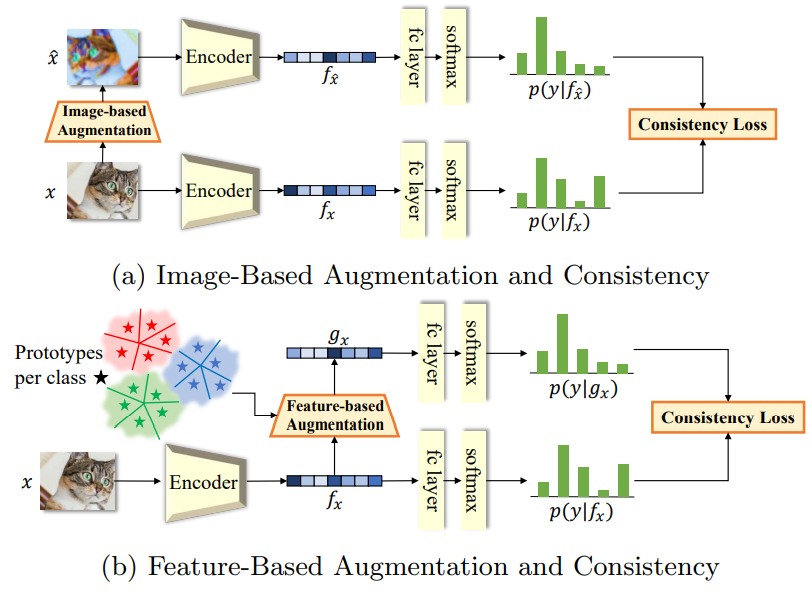}}
\caption{An overview of featMatch augmentation applied on images and features. Image is taken from \cite{chu2020feature}.}
\label{fig:featMatch}
\end{figure}

\item{\textbf{Dataset Augmentation in Feature Space:}}
This work~\cite{devries2017dataset} first used encoder-decoder to learn representation, then on representation apply different transformations such as adding noise, interpolating, or extrapolating. The proposed method has shown performance improvement on both static and sequential data. Moreover, a demonstration of this augmentation is shown in figure~\ref{fig:dafs}.

\begin{figure}[htbp]
{\includegraphics[width=0.5\textwidth]{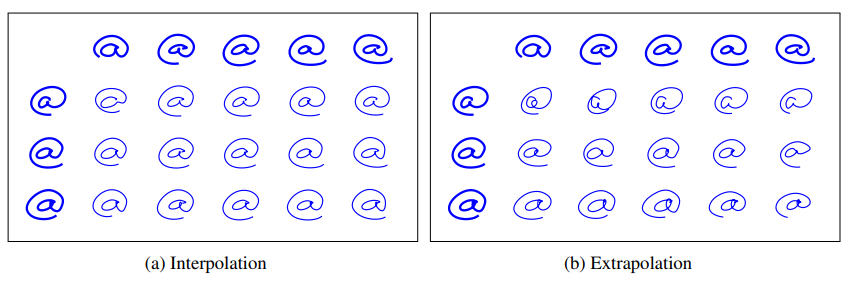}}

\caption{Overview of interpolation and extrapolation between handwritten characters. Original characters are shown in bold. Image is taken from~\cite{devries2017dataset}.}
\label{fig:dafs}
\end{figure}

\item {\textbf{Feature Space Augmentation for Long-Tailed Data :}}
This paper~\cite{chu2020feature} proposed the novel data augmentation in feature space to address the long-tailed issue and uplift the under-represented class samples. The proposed approach first separates class-specific features into generic and specific features with the help of class activation maps.  Under-represented class samples are generated by injecting class-specific features of under-represented classes with class-generic features from other confusing classes. It enables diverse data and also deals with the problem of under-represented class samples. It has shown SOTA performance on different datasets. It is demonstrated in figure~\ref{fig:fsaltd}.

\begin{figure}[htbp]
{\includegraphics[width=0.5\textwidth]{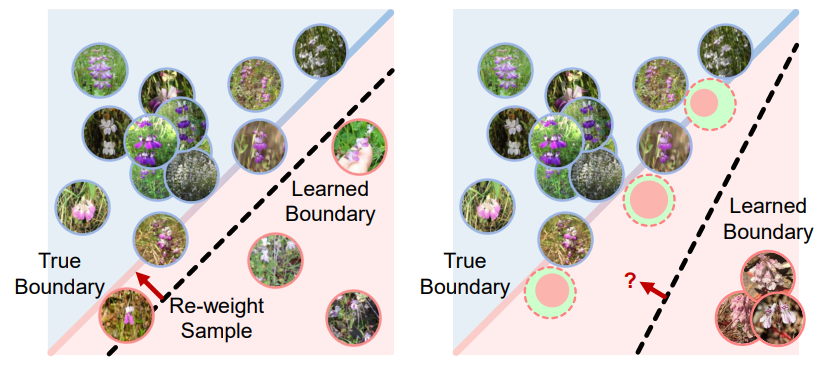}}
\caption{Left: limited but well-spread data. Right: Without sufficient data. Image is taken from~\cite{chu2020feature}.}
\label{fig:fsaltd}
\end{figure}

\item {\textbf{Adversarial Feature Augmentation for Unsupervised Domain Adaptation: }}
Generative Adversarial Networks (GANs) showed promising results in unsupervised domain adaptation to learn target domain features indistinguishable from the source domain. This work~\cite{volpi2018adversarial} extends GAN  by contributing: i) it forces feature extractor to be domain-invariant ii) To train it via data augmentation in feature space, named feature augmentation. This work explores data augmentation at the feature level with GAN.

\item {\textbf{Understanding data augmentation for classification: when to warp? :}}
This paper~\cite{wong2016understanding} investigates the data augmentation advantages on image space and feature space during training. It proposed two approaches i) data warping which generates extra samples in image space using data augmentations and ii) synthetic over-sampling, which generates samples in feature space. It also suggests that it is possible to apply general data augmentation techniques in feature space if reasonable data augmentations for data are known.

\end{enumerate}

\subsubsection{\textbf{Neural Style Transfer}}
\begin{figure}[htbp]
{\includegraphics[width=0.5\textwidth]{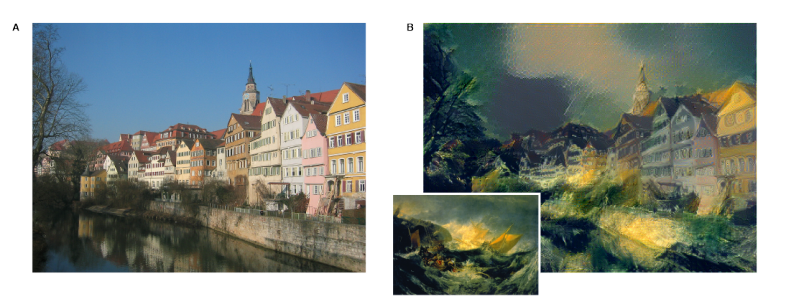}}
\caption{Overview of the styled image by the neural algorithm. Image is from~\cite{gatys2015neural}.}
\label{fig:nafst}
\end{figure}

It is another category of data augmentation, which can transfer the artist style of one image to another without changing semantics at a high level. It brings more variety to the training set. The main objective of this neural style transfer is to generate a third image from two images, where one image provides texture content and another provides high-level semantic content. We explore some of the SOTA augmentations for the sub-category. 
\begin{enumerate}[label=(\roman*)]
\item {\textbf{STaDA: Style Transfer as Data Augmentation : }}
This work~\cite{zheng2019stada} thoroughly evaluated different SOTA neural style transfer algorithms as data augmentation for image classification tasks. It shows significant performance gain on Caltech 101~\cite{fei2006one} and Caltech 256~\cite{griffin2007caltech} datasets. Furthermore, it also combines neural style transfer algorithms with conventional data augmentation methods.  A sample of this augmentation is shown in figure~\ref{fig:stada}.

\begin{figure}[htbp]
{\includegraphics[width=0.5\textwidth]{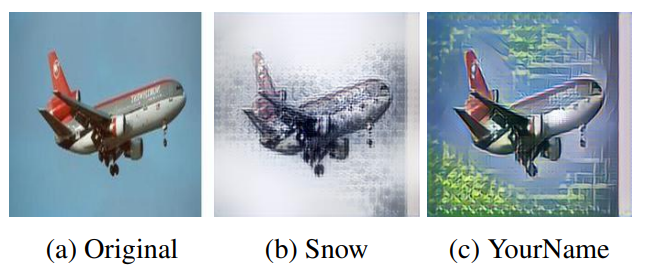}}
\caption{Overview of the original image and two stylized images by STaDA. Image is taken from~\cite{zheng2019stada}.}
\label{fig:stada}
\end{figure}

\item {\textbf{Style Augmentation: Data Augmentation via Style Randomization: }}
This work~\cite{jackson2019style} proposed a novel data augmentation named style augmentation (SA)  based on style neural transfer. SA randomizes the color, contrast, and texture while maintaining the shape and semantic content during the training. This is done by picking an arbitrary style transfer network for randomizing the style and by getting the target style from multivariate normal distribution embedding. It improves performance in three different tasks: classification, regression, and domain adaptation. The style augmentation sample is shown in figure~\ref{fig:style_aug}.

\begin{figure}[htbp]
{\includegraphics[width=0.5\textwidth]{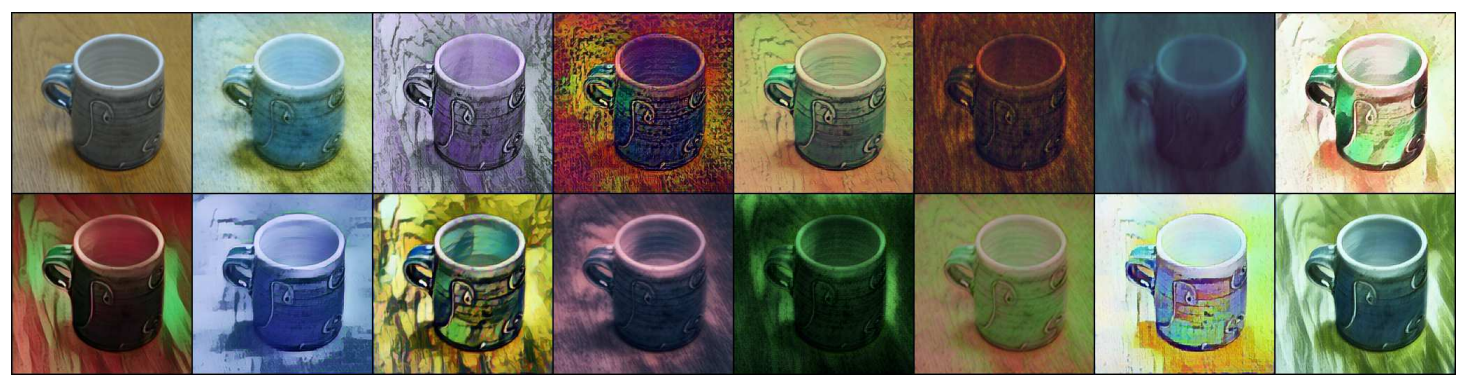}}
\caption{Overview of Style augmentation applied to an image. The shape is preserved but the style, including color, texture, and contrast is randomized. Image is from~\cite{jackson2019style}.}
\label{fig:style_aug}
\end{figure}

\item {\textbf{StyPath: Style-Transfer Data Augmentation for Robust Histology Image Classification:}}
This paper~\cite{cicalese2020stypath} proposes a novel pipeline for Antibody Mediated Rejection (AMR) classification in kidneys based on StyPath data augmentation. StyPath is data augmentation that transfers style intending to reduce bias. The proposed augmentation is much faster than SOTA augmentations for AMR classification.  Some samples are shown in figure~\ref{fig:stypath}.

\begin{figure}[htbp]
{\includegraphics[width=0.5\textwidth]{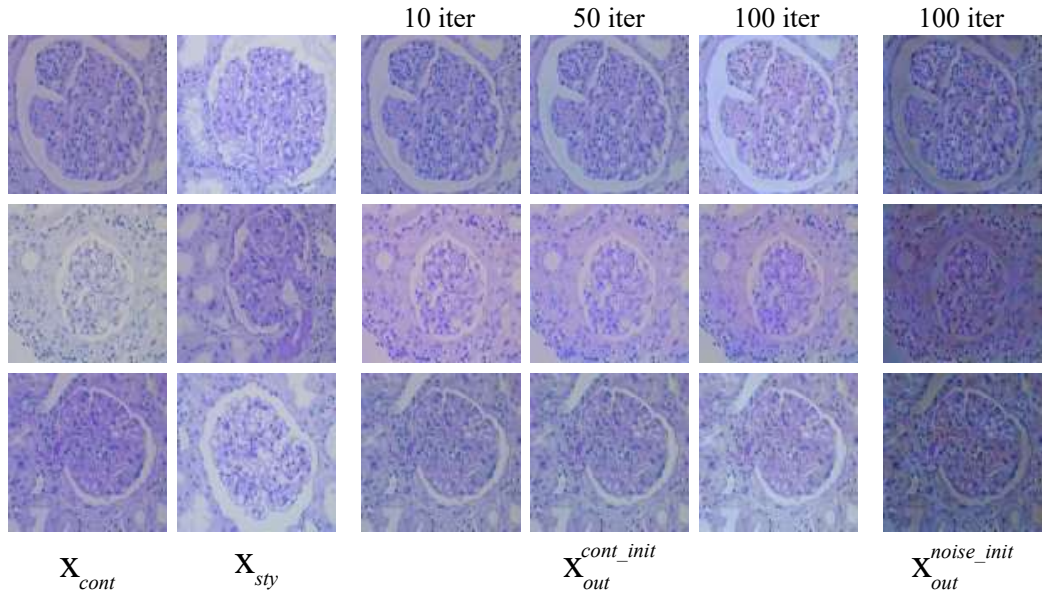}}
\caption{Comparison of content and random initialization. Authors observe that output images initialized as the noise appeared distorted and discolored and failed to retain the content fidelity.  Image is from~\cite{cicalese2020stypath}.}
\label{fig:stypath}
\end{figure}

\item {\textbf{A Neural Algorithm of Artistic Style : }}
This work~\cite{gatys2015neural} introduces an artificial system (AS) based on a deep neural network that generates artistic images of high perceptual quality. AS creates neural embedding then it uses the embedding to separate the style and content of the image and then recombines the content and style of target images to generate the artistic image. The sample is shown in figure~\ref{fig:nafst}

\item {\textbf{Neural Style Transfer as Data Augmentation for Improving COVID-19 Diagnosis Classification :}}
This work~\cite{hernandez2021neural} shows the effectiveness of a cycle GAN, which is mostly used for neural style transfer, augments COVID-19 negative x-ray image to convert into a positive COVID image to balance the dataset and also to increase the diversity of the dataset. It shows that augmenting the images with cycle GAN can improve performance over several different CNN architectures.  A sample of this augmentation is shown in figure ~\ref{fig:NSDACOVID}.

\begin{figure}[htbp]
{\includegraphics[width=0.5\textwidth]{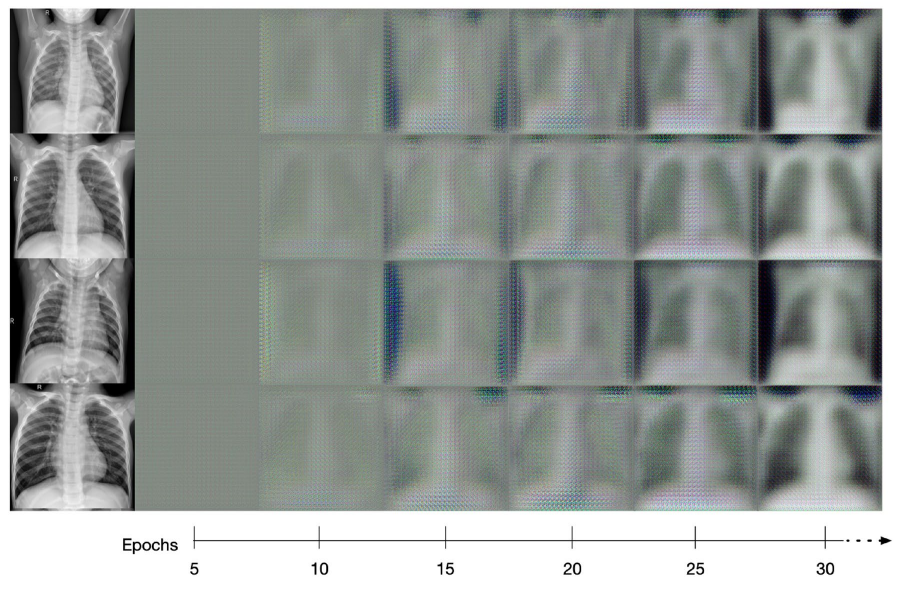}}
\caption{Overview of generating synthetic COVID images from the healthy category. As the no of epochs grows the quality of the synthetic images improves. An example is from~\cite{hernandez2021neural}.}
\label{fig:NSDACOVID}
\end{figure}

\end{enumerate}
\begin{table*}[ht]
\centering
\begin{tabular}{@{\extracolsep{4pt}}llccccccc}
\toprule   
{} & \multicolumn{4}{c}{Accuracies} \\
 \cmidrule{2-5} 

 Method & CIFAR10 & CIFAR10+ & CIFAR100 & CIFAR100+ \\
 \midrule ResNet-18 (Baseline) & 89.37 & 95.28 & 63.32 & 77.54 \\
ResNet-18 + CutOut & 90.69 & 96.25 & 65.02 & 80.58 \\
ResNet-18 + Random Erasing & 95.28 & 95.32 & - & - \\
ResNet-18 + CutMix & 90.56 & 96.22 & 65.58 & 80.58 \\
ResNet-18 + SaliencyMix & 92.41 & 96.35 & 71.27 & 80.71 \\
ResNet-18 + GridMask   & 95.28 & 96.54 & - & - \\
\midrule ResNet-50 (Baseline) & 87.86 & 95.02 & 63.52 & 78.42 \\
ResNet-50 + CutOut & 91.16 & 96.14 & 67.03 & 78.62 \\
ResNet-50 + CutMix & 90.84 & 96.39 & 68.35 & 81.28 \\
ResNet-50 + SaliencyMix & 93.19 & 96.54 & 75.11 & 81.43 \\
\midrule WideResNet-28-10 (Baseline)~\cite{uddin2020saliencymix}  & 93.03 & 96.13 & 73.94 & 81.20 \\ 
 WideResNet-28-10 + CutOut\cite{devries2017improved}   & 94.46 & 96.92 & 76.06 & 81.59 \\ 
WideResNet-28-10 + Random Erasing   & 96.2 & 96.92 & 81.59 & 82.27 \\  
WideResNet-28-10 + GridMask   & 96.13 & 97.24 & - & - \\
WideResNet-28-10 + CutMix   & 94.82 & 97.13 & 76.79 & 83.34 \\
WideResNet-28-10 + PuzzleMix   & - & - & - & 83.77 \\
WideResNet-28-10 + SaliencyMix   & 95.96 & 97.24 & 80.55 & 83.44 \\

\bottomrule
\multicolumn{4}{l}{Note: $+$ sign after dataset name shows that traditional data augmentation methods have been used}
\end{tabular}
\label{classification_results1}
\caption{Baseline performance comparison of various augmentation on CIFAR10 and CIFAR100 datasets.} 
\end{table*}

\begin{table*}[ht]
\centering

\begin{tabular}{@{\extracolsep{4pt}}llccccccc}
\toprule   
{} & \multicolumn{2}{c}{CIFAR-10}  & \multicolumn{2}{c}{CIFAR-100} & \multicolumn{2}{c}{ImageNet}\\
 \cmidrule{2-3} 
 \cmidrule{4-5} 
 \cmidrule{6-7} 
 Augmentation & Accuracy (\%) & Model & Accuracy (\%) & Model & Accuracy (\%) & Model \\ 
\midrule
Cutout\cite{devries2017improved} & 97.04 & WRN-28-10 & 81.59 & WRN-28-10 & 77.1 & ResNet-50  \\ 
Random Erasing\cite{zhong2020random} & 96.92 & WRN-28-10 & 82.27 & WRN-28-10 & - & - \\ 
Hide-and-Seek\cite{singh2018hide} & 95.53 & ResNet-110 & 78.13 & ResNet-110 & 77.20 & ResNet-50 \\
GridMask\cite{chen2020gridmask} & 97.24 & WRN-28-10 & - & - & 77.9 & ResNet-50 \\
LocalAugment\cite{kim2021local} & - & - & 95.92 & WRN-22-10 & 76.87 & ResNet-50 \\
SalfMix\cite{choi2021salfmix} & 96.62 & PreActResNet-101 & 80.11 & PreActResNet-101 & - & - \\
\bottomrule
\end{tabular} 
\caption{Performance comparison of the various image erasing and image mixing augmentations for image classification problems. WRN and SS stand for WideResNet and Shake-Shake, respectively.}
\label{classification_results2} 
\end{table*}

\begin{table*}[ht]
\centering
\caption{Comparison on CIFAR-10 and SVHN. The number represents error rates across three runs.}
\begin{tabular}{@{\extracolsep{4pt}}llccccccc}
\toprule   
{} & \multicolumn{4}{c}{CIFAR-10} & \multicolumn{3}{c}{SVHN} \\
 \cmidrule{2-5}
 \cmidrule{6-9}

 Method & 40 labels & 250 labels & 1,000 labels & 4,000 labels & 40 labels & 250 labels & 1,000 labels & 4,000 labels \\
 \midrule 
 VAT\cite{miyato2018virtual} &-& 36.03 $\pm$ 2.82 & 18.64 $\pm$ 0.40 & 11.05 $\pm$ 0.31 & - & 8.41 $\pm$ 1.01 &  5.98 $\pm$ 0.21 & 4.20 $\pm$ 0.15 \\
 Mean Teacher~\cite{tarvainen2017mean} &-  & 47.32 $\pm$ 4.71 & 17.32$\pm$4.00  & 10.36$\pm$0.25  & - & 6.45$\pm$2.43  & 3.75$\pm$.10   & 3.39$\pm$0.11   \\
MixMatch~\cite{berthelot2019mixmatch}  &47.54$\pm$11.50     & 11.08$\pm$.87 & 7.75$\pm$.32   & 6.24$\pm$.06   & 42.55$\pm$14.53 & 3.78$\pm$.26  & 3.27$\pm$.31   & 2.89$\pm$.06   \\
ReMixMatch~\cite{berthelot2019remixmatch} &19.10$\pm$9.64     & 6.27$\pm$0.34  & 5.73$\pm$0.16   & 5.14$\pm$0.04 & 3.34$\pm$0.20  & 3.10$\pm$0.50  & 2.83$\pm$0.30   & 2.42$\pm$0.09   \\
UDA &29.05$\pm$5.93 & 8.76$\pm$ 0.90  & 5.87$\pm$ 0.13   & 5.29$\pm$ 0.25 & 52.63$\pm$20.51   & 2.76$\pm$ 0.17  & 2.55$\pm$ 0.09   & 2.47$\pm$ 0.15   \\
SSL with Memory~\cite{chen2018semi} & - & - &  - &  11.9$\pm$0.22 & - & 8.83 & 4.21 & - \\
\bottomrule
\end{tabular} 
\label{SSL1}
\end{table*}

\begin{table*}[ht]
\centering
\caption{Comparison on CIFAR-100 and mini-ImageNet. The number represents error rates across two runs.} 
\begin{tabular}{@{\extracolsep{4pt}}llccccccc}
\toprule   
{} & \multicolumn{3}{c}{CIFAR-100} & \multicolumn{2}{c}{mini-ImageNet} \\
 \cmidrule{2-4}
 \cmidrule{5-6}
  Method & 400 labels & 4,000 labels & 10,000 labels & 4,000 labels & 10,000 labels \\ \midrule 
  II-model\cite{sajjadi2016regularization} & - & - & 39.19$\pm$ 0.36 & - & - \\
SNTG\cite{luo2018smooth} & - & - & 37.97$\pm$ 0.29 & - & - \\
SSL with Memory\cite{chen2018semi} & - & - & 34.51$\pm$ 0.61 & - & - \\
Deep Co-Training\cite{qiao2018deep} & - & - & 34.63$\pm$ 0.14 & - & - \\
Weight Averaging\cite{athiwaratkun2018improving} & - & - & 33.62$\pm$ 0.54 & - & - \\
Mean Teacher\cite{tarvainen2017mean} & - & 45.36 $\pm$0.49 & 36.08$\pm$ 0.51 & 72.51$\pm$ 0.22 & 57.55 $\pm$ 1.11 \\
Label Propagation\cite{iscen2019label} & - & 43.73 $\pm$0.20 & 35.92 $\pm$0.47 & 70.29$\pm$ 0.81 & 57.58 $\pm$1.47 \\
PLCB\cite{arazo2020pseudo} & - & 37.55 $\pm$1.09 & 32.15 $\pm$0.50 & 56.49 $\pm$0.51 & 46.08 $\pm$ 0.11 \\
FeatMatch & - & 31.06 $\pm$ 0.41 & 26.83 $\pm$ 0.04 & 39.05 0.06 & 34.79$\pm$0.22 \\
MixMatch & 67.61$\pm$1.32 & - & 28.31$\pm$0.33 & - & - \\
UDA & 59.28$\pm$0.88 & - & 24.50$\pm$0.25 & - & - \\
ReMixMatch & 44.28$\pm$2.06 & - &  23.03$\pm$0.56 & - & - \\
FixMatch & 48.85$\pm$1.75 & - & 22.60$\pm$0.12 & - & - \\
 \bottomrule
\end{tabular}
\label{SSL2}
\end{table*}

 \begin{table*}[ht]
\centering
\caption{Comparison of test error rates on CIFAR-10 \& SVHN using WideResNet-28 and CNN-13.} 
\begin{tabular}{@{\extracolsep{4pt}}llccccccc}
\toprule   
  Approach & Method & CIFAR-10 ($N_{l}$=4000) & SVHN($N_{l}$=1000)  \\\midrule 
  WideResNet-28 \\ \midrule 
  & Supervised & 20.26 $\pm$ 0.38 & 12.83 $\pm$ 0.47 \\
Pseudo & PL \cite{lee2013pseudo} & 17.78 $\pm$ 0.57 & 7.62 $\pm$ 0.29 \\
Labeling & PL-CB \cite{arazo2020pseudo} & 6.28 $\pm$ 0.3 & - \\
 & II Model \cite{laine2016temporal} & 16.37 $\pm$ 0.63 & 7.19 $\pm$ 0.27 \\
 & Mean Teacher \cite{tarvainen2017mean} & 15.87 $\pm$ 0.28 & 5.65 + 0.47 \\
 & VAT \cite{miyato2018virtual} & 13.86 $\pm$ 0.27 & 5.63 $\pm$ 0.20 \\
\bottomrule
\end{tabular}
\label{SSL3}
\end{table*}

\begin{table*}[]

\begin{tabular}{llllllllll}
&   &   &  &    &  &  &    &  \\ \hline
\multicolumn{1}{|l|}{\textbf{Method}} & \multicolumn{1}{|l|}{\textbf{Detector}} &   \multicolumn{1}{|l|}{\textbf{BackBone}} & \multicolumn{1}{|l|}{\textbf{AP}}&\multicolumn{1}{|l|}{\textbf{$AP_50$}} &\multicolumn{1}{|l|}{\textbf{$AP_75$}} &   \multicolumn{1}{|l|}{\textbf{$AP_s$}} & \multicolumn{1}{|l|}{\textbf{$AP_m$}}&\multicolumn{1}{|l|}{\textbf{$AP_l$}} \\ 

\hline  
\multicolumn{1}{|l|}{Hand-crafted:}                 & \multicolumn{8}{l|}{}    \\ \hline                                                                 
\multicolumn{1}{|l|}{Dropblock~\cite{ghiasi2018dropblock}}  & \multicolumn{1}{l|}{RetinaNet}    & \multicolumn{1}{l|}{ResNet-50} & \multicolumn{1}{l|}{38.4} & \multicolumn{1}{l|}{56.4} & \multicolumn{1}{l|}{41.2}      & \multicolumn{1}{l|}{$-$}       & \multicolumn{1}{l|}{$-$} & \multicolumn{1}{l|}{$-$}      \\ \hline
\multicolumn{1}{|l|}{AutoAugment+color Ops~\cite{zoph2020learning}} & \multicolumn{1}{l|}{RetinaNet}    & \multicolumn{1}{l|}{ResNet-50} & \multicolumn{1}{l|}{37.5} & \multicolumn{1}{l|}{-} & \multicolumn{1}{l|}{-}      & \multicolumn{1}{l|}{$-$}       & \multicolumn{1}{l|}{$-$} & \multicolumn{1}{l|}{$-$}\\ \hline
\multicolumn{1}{|l|}{geometric Ops~\cite{zoph2020learning}} & \multicolumn{1}{l|}{RetinaNet} & \multicolumn{1}{l|}{ResNet-50} & \multicolumn{1}{l|}{38.6} & \multicolumn{1}{l|}{-} & \multicolumn{1}{l|}{-}      & \multicolumn{1}{l|}{$-$} & \multicolumn{1}{l|}{$-$} & \multicolumn{1}{l|}{$-$}\\ \hline
\multicolumn{1}{|l|}{bbox-only Ops~\cite{zoph2020learning}} & \multicolumn{1}{l|}{RetinaNet} & \multicolumn{1}{l|}{ResNet-50} & \multicolumn{1}{l|}{39.0} & \multicolumn{1}{l|}{-} & \multicolumn{1}{l|}{-} & \multicolumn{1}{l|}{$-$} & \multicolumn{1}{l|}{$-$} & \multicolumn{1}{l|}{$-$}\\
\hline
\end{tabular}
\label{tab:object_detection_results_PASCAL_VOC_1} 
\caption{Data augmentation effect on different object detection methods using PASCAL VOC dataset }
\end{table*}

\begin{sidewaystable*}[]
\hspace*{-38pt}
\small
\begin{tabular}{@{}|l|l|l|l|l|l|l|l|l|l|l|l|l|l|l|l|l|l|l|l|l|l|l|@{}}
\hline
Method         & TSet & mAP  & aero & bike & bird & boat & bottle & bus  & car  & cat  & chair & cow  & table & dog  & horse & mbike & person & plant & sheep & sofa & train & tv   \\
\hline
FRCN\cite{girshick2015fast}   & 7        & 66.9 & 74.5 & 78.3 & 69.2 & 53.2 & 36.6   & 77.3 & 78.2 & 82.0 & 40.7  & 72.7 & 67.9  & 79.6 & 79.2  & 73.0  & 69.0  & 30.1  & 65.4  & 70.2 & 75.8  & 65.8 \\
\hline
FRCN*\cite{wang2017fast} & 7        & 69.1 & 75.4 & 80.8 & 67.3 & 59.9 & 37.6   & 81.9 & 80.0 & 84.5 & 50.0  & 77.1 & 68.2  & 81.0 & 82.5  & 74.3  & 69.9  & 28.4  & 71.1  & 70.2 & 75.8  & 66.6 \\
\hline
ASDN\cite{wang2017fast}  & 7        & 71.0 & 74.4 & 81.3 & 67.6 & 57.0 & 46.6   & 81.0 & 79.3 & 86.0 & 52.9  & 75.9 & 73.7  & 82.6 & 83.2  & 77.7  & 72.7  & 37.4  & 66.3  & 71.2 & 78.2  & 74.3 \\ \hline
IRE     & 7        & 70.5 & 75.9 & 78.9 & 69.0 & 57.7 & 46.4   & 81.7 & 79.5 & 82.9 & 49.3  & 76.9 & 67.9  & 81.5 & 83.3  & 76.7  & 73.2  & 40.7  & 72.8  & 66.9 & 75.4  & 74.2 \\ \hline
ORE     & 7        & 71.0 & 75.1 & 79.8 & 69.7 & 60.8 & 46.0   & 80.4 & 79.0 & 83.8 & 51.6  & 76.2 & 67.8  & 81.2 & 83.7  & 76.8  & 73.8  & 43.1  & 70.8  & 67.4 & 78.3  & 75.6 \\ \hline
I+ORE   & 7        & 71.5 & 76.1 & 81.6 & 69.5 & 60.1 & 45.6   & 82.2 & 79.2 & 84.5 & 52.5  & 78.7 & 71.6  & 80.4 & 83.3  & 76.7  & 73.9  & 39.4  & 68.9  & 69.8 & 79.2  & 77.4 \\ \hline
FRCN \cite{girshick2015fast}   & 7+12     & 70.0 & 77.0 & 78.1 & 69.3 & 59.4 & 38.3   & 81.6 & 78.6 & 86.7 & 42.8  & 78.8 & 68.9  & 84.7 & 82.0  & 76.6  & 69.9  & 31.8  & 70.1  & 74.8 & 80.4  & 70.4 \\
\bottomrule
\end{tabular}
\caption{VOC 2007 test detection average precision (\%). FRCN* refers to FRCN with training schedule in \cite{wang2017fast} and SD refers to synthetic data}
\label{tab:object_detection_pascal_voc}.
\end{sidewaystable*}



\begin{table}[]
\centering

\begin{tabular}{@{}|l|l|l|l|@{}}
\hline
Model                & mAP         & AP50        & AP75        \\ \hline
EfficientDet-D0      & 55.6        & 77.6        & 61.4        \\ \hline
+ AutoAugment        & 55.7 (+0.1) & 77.7 (+0.1) & 61.8 (+0.4) \\ \hline
+ Det-AdvProp & 55.9 (+0.3) & 77.9 (+0.3) & 62.0 (+0.6) \\ \hline
EfficientDet-D1      & 60.8        & 82.0        & 66.7        \\ \hline
+ AutoAugment        & 61.0 (+0.2) & 82.2 (+0.2) & 67.2 (+0.5) \\ \hline
+ Det-AdvProp & 61.2 (+0.4) & 82.3 (+0.3) & 67.4 (+0.7) \\ \hline
EfficientDet-D2      & 63.3        & 83.6        & 69.3        \\ \hline
+ AutoAugment        & 62.7 (-0.6) & 83.3 (-0.3) & 69.2 (-0.1) \\ \hline
+ Det-AdvProp & 63.5 (+0.2) & 83.8 (+0.2) & 69.7 (+0.4) \\ \hline
EfficientDet-D3      & 65.7        & 85.3        & 71.8        \\ \hline
+ AutoAugment        & 65.2 (-0.5) & 85.1 (-0.2) & 71.3 (-0.5) \\ \hline
+ Det-AdvProp & 66.2 (+0.5) & 85.9 (+0.6) & 72.5 (+0.7) \\ \hline
EfficientDet-D4      & 67.0        & 86.0        & 73.0        \\ \hline
+ AutoAugment        & 67.0 (+0.0) & 86.3 (+0.3) & 73.5 (+0.5) \\ \hline
+ Det-AdvProp & 67.5 (+0.5) & 86.6 (+0.6) & 74.0 (+1.0) \\ \hline
EfficientDet-D5      & 67.4        & 86.9        & 73.8        \\ \hline
+ AutoAugment        & 67.6 (+0.2) & 87.2 (+0.3) & 74.2 (+0.4) \\ \hline
+ Det-AdvProp & 68.2 (+0.8) & 87.6 (+0.7) & 74.7 (+0.9) \\ \hline
\end{tabular}
\caption{Results on PASCAL VOC 2012. The proposed DetAdvProp gives the highest score on every model and metric. It largely outperforms AutoAugment\cite{cubuk2019autoaugment} when facing domain shift.}
\label{tab:object_detection_pascal_voc_2012}
\end{table}

\begin{table*}[ht]
\centering
\begin{tabular}{|l|l|l|l|l|l|l|}
\hline
\textbf{Method}  & \textbf{Model} & \textbf{1/8} & \textbf{1/4} & \textbf{1/2} & \textbf{7/8} & \textbf{Full} \\ \hline
SDA~\cite{yuan2021simple}           & DeepLabV3Plus  & 74.1         & -            & -            & -            & -             \\ \hline
SDA + DSBN~\cite{yuan2021simple}      & DeepLabV3Plus  & 69.5         & -            & -            & -            & -             \\ \hline
SDA~\cite{yuan2021simple}              & DeepLabV3Plus  & -            & -            & -            & -            & 78.7          \\ \hline
SDA + DSBN~\cite{yuan2021simple}       & DeepLabV3Plus  & -            & -            & -            & -            & 79.2          \\ \hline
SDA~\cite{yuan2021simple}             & DeepLabV3Plus  & -            & -            & -            & 71.4         & -             \\ \hline
SDA + DSBN~\cite{yuan2021simple}       & DeepLabV3Plus  & -            & -            & -            & 72.5         & -             \\ \hline
AdvSemi~\cite{hung2018adversarial}         & DeepLabV2      & 58.8         & 62.3         & 65.7         & -            & 66.0          \\
\hline

\end{tabular}
\caption{Results of Performance (mIoU) on Cityscapes validation set}
\label{cityScapes}
\end{table*}

\begin{table*}[htp]
\centering
\begin{tabular}{|l|l|l|l|l|l|l|l|}
\hline
\textbf{Method}                              & \textbf{Model}  & \textbf{1/100} & \textbf{1/50} & \textbf{1/20} & \textbf{1/8} & \textbf{1/4} & \textbf{Full} \\ \hline
GANSeg~\cite{souly2017semi}           & VGG16          & -            & -            & -            & -            & 64.1    &      \\ \hline
AdvSemSeg~\cite{hung2018adversarial}       & ResNet-101     & -            & -            & -            & -            & 68.4          & \\ \hline
CCT~\cite{ouali2020semi}            & ResNet-50      & -            & -            & -            & -            & 69.4          & \\ \hline
PseudoSeg~\cite{zou2020pseudoseg}       & ResNet-101     & -            & -            & -            & -            & 73.2        &  \\ \hline
DSBN~\cite{yuan2021simple}            & ResNet-101     & -            & -            & -            & -            & 75.0    &      \\ \hline
DSBN~\cite{yuan2021simple}            & Xception-65    & -            & -            & -            & -            & 79.3      &    \\ \hline
Fully supervised~\cite{yuan2021simple} & ResNet-101     & -            & -            & -            & -            & 78.3      &    \\
\hline
\end{tabular}
\caption{Results of Performance mean intersection over union (mIoU) on the Pascal VOC 2012 validation set}
\label{PASVOC2012}
\end{table*}


\section{Results}
In this section, we provide the detailed result for various Computer Vision tasks such as image classification, object detection, and semantic segmentation. The main purpose is to show the effect of the data augmentation in CV different tasks and to do so, we compile results from various SOTA data augmentation works.

\subsection{Image Classification}
In this section, we present the result of several SOTA data augmentation methods for supervised learning and semi-supervised learning. Both are discussed below:  
 \subsubsection{supervised learning results}
 In supervised learning, we have data on a large quantity that is fully labeled and we use this data to train the neural network (NN) model.  In this section, we compile and  compare the results from several SOTA data augmentation methods and put them in two different tables as shown in table~\ref{classification_results1} and table~\ref{classification_results2}. In table~\ref{classification_results1} results, $^+$ sign shows traditional data augmentations such as flipping, rotating, and cropping, have been used along with the SOTA augmentation methods. The used datasets are CIFAR10~\cite{krizhevsky2009learning}, CIFAR100~\cite{krizhevsky2009learning} and ImageNet~\cite{5206848}, and the used networks are wideresnet flavours~\cite{he2016deep}, pyramid network flavours and several popular resnet flavours~\cite{he2016deep}. Accuracy is the evaluation metric used to compare the different algorithms used. The higher the accuracy, the better. As it can be in table~\ref{classification_results1} and  table~\ref{classification_results2}, each data augmentation has significantly improved the accuracy.

\subsubsection{Semi-supervised learning}

Semi-supervised learning (SSL) is when we have a limited labeled data but unlabeled data is available on the large scale. Labeling the unlabeled data is tedious, time-consuming, and costly~\cite {kumar2021binary, xie2010semi}. To avoid these issues,  SSL is used. There are several techniques of SSL, but recently, data augmentation is employed with the limited labeled data to increase the diversity of the data. Data augmentation with SSL has increased the performance on different datasets and NN architectures.  The used dataset are CIFAR10, CIFAR100, SVHn~\cite{netzer2011reading} and Mini-ImageNet. Several SSL techniques are used such as pseudoLabel, SSL with memory, label propagation, mean teacher, etc.  We compile the results from many SOTA SSL methods with data augmentation and present them in this work. The effect of the data augmentation has also been shown with the different number of samples in SSL as shown in table~\ref{SSL1}, table~\ref{SSL2}, and table~\ref{SSL3}.

\subsection{Object detection}
In this section, we discuss the effectiveness of various image data augmentation techniques on the frequently used COCO2017~\cite{lin2014microsoft}, PASCAL VOC~\cite{everingham2009pascal}, VOC 2007~\cite{pascal-voc-2007}, and VOC 2012~\cite{pascal-voc-2012} datasets, which are commonly used for object detection tasks. We compile results from various SOTA data augmentation methods and put them in three different tables as shown in the table \ref{tab:object_detection_results_PASCAL_VOC_1}, \ref{tab:object_detection_pascal_voc}, and \ref{tab:object_detection_pascal_voc_2012}. FRCNN along with synthetic data gives the best mAP accuracy on VOC 2007 dataset as shown in
table~\ref{tab:object_detection_pascal_voc}. Several classical and automatic data augmentation methods have shown promising performance using different SOTA models on the PASCAL VOC dataset as shown in table \ref{tab:object_detection_results_PASCAL_VOC_1}.
The DetAdvProp achieves the highest score outperforming AutoAugment~\cite{cubuk2019autoaugment} on PASCAL VOC 2012 dataset as shown in the table \ref{tab:object_detection_pascal_voc_2012}. The scores are in terms of mean average precision (mAP), average precision (AP) at the intersection over union (IOU) of 0.5 (AP50),  and AP at IOU of 0.75 (AP75) metrics.  

\subsection{Semantic Segmentation}
This subsection includes semantic segmentation results on PASCAL VOC  and CITYSCAPES datasets, most frequently used in several research papers. In table~\ref{cityScapes} and table~\ref{PASVOC2012}, we compiled the effectiveness of validation set results on the different datasets with the effect of SOTA data augmentations on the semantic segmentation task.  The results are reported in the term of mean intersection over union (mIoU) as the accuracy on the Cityscape dataset and PASCAL VOC dataset as shown in table~\ref{cityScapes} and table~\ref{PASVOC2012}, respectively. We found performance gains on a few metrics such as mIoU and mAP, with several semantic segmentation models:
 deeplabv3+~\cite{yuan2021simple}, DeepLab-v2~\cite{olsson2021classmix}, Xception-65~\cite{yuan2021simple}, ExFuse~\cite{zhang2018exfuse} and Eff-L2~\cite{zoph2020rethinking} .  It has been observed that incorporating data augmentation techniques can enhance the performance of semantic segmentation models. Notably, advanced image data augmentation methods have demonstrated greater improvements in performance compared to traditional techniques. Table~\ref{cityScapes} and table~\ref{PASVOC2012} provide evidence of this improvement.  The traditional data augmentations including rotation, scaling, flipping, and shifting~\cite{zhang2021objectaug}.

\section{Discussion and future directions}\label{discussion}
\subsection{\textbf{Current approaches}} It is proven that if we provide more data to the model, it improves model performance~\cite{sun2017revisiting,halevy2009unreasonable}. A few current tendencies are discussed by Xu et al. ~\cite{xu2022comprehensive}. Among these, one way is to collect the data and label it manually, but it is not an efficient way to do this. Another efficient way is to apply data augmentation, the more data augmentations we apply, the better improvement  we get in terms of performance but to a certain extent. Currently, image mixing methods and autoaugment methods are successful for image classification tasks, scale aware based auto augment methods are showing promising results in detection tasks and semantic segmentation tasks. But these data augmentation performances can vary with the number of data augmentation applied, as it is known that the combined data augmentation methods show better performance than single one ~\cite{yang2022image,pawara2017data}.    
\subsection{\textbf{Theoretical aspects}} There is no theoretical support available to explain why specific augmentation is improving performance and which sample(s) should be augmented, as the same aspect has been discussed by Yang et al~\cite{yang2022image} and Shorten et al~\cite{shorten2019survey}. Like in random erasing, we randomly erase the region of the image - sometime may erase discriminating features, and the erased image makes no sense to a human. But the reason behind performance improvement is still unknown, which is another open challenge. Most of the time, we find the optimal parameters of the augmentation through an extensive number of experiments or we choose data augmentation based on our experience. But there should be a mechanism for choosing the data augmentation with theoretical support considering model architecture and dataset size. Researching the theoretical aspect is another open challenge for the research community.  
\subsection{\textbf{Optimal number of samples generation}} It is a known fact, as we increase data size, it improves the performance ~\cite{sun2017revisiting,halevy2009unreasonable,yang2022image,shorten2019survey} but it is not a case - increasing the number of samples will not improve performance after a certain number of samples ~\cite{kumar2021class}. What is the optimal number of samples to be generated, depending on the model architecture and dataset size, is a challenging aspect to be explored. Currently, researchers perform many experiments to find the optimal number of sample generation~\cite{kumar2021class}. But it is not feasible way as it requires time and computational cost. Can we devise a mechanism to find an optimal number of samples, which is an open research challenge?      
\subsection{\textbf{Selection of data augmentation based on model architecture and dataset}} Data augmentation selection depends on the nature of the dataset and model architecture. Like on MNIST~\cite{deng2012mnist} dataset, geometric transformations are not safe such as rotation on 6 and 9 digits will no longer preserve the label information. For densely parameterized CNN, it is easy to overfit weakly augmented datasets, and for shallow parameterized CNN, it may break generalization capability with data augmentation. It suggests, while selecting the data augmentation, the nature of the dataset and model architecture should be taken into account.  Currently, numerous experiments are performed to find model architecture and suitable data augmentation for a specific dataset. Devising a systematic approach to select the data augmentation based on dataset and model architecture is another gap to be filled.   
\subsection{\textbf{Augmentations for spaces}} Most of the data augmentation approaches have been explored on the image level - data space. Very few research works have explored data on feature level - feature space. The challenge here arises, in which space should we apply data augmentation, data space, or feature space? It is another interesting aspect that can be explored.  For the current approaches, it seems like it depends on the dataset, model architecture, and task. Currently, approaches are conducting experiments in data space and feature space and then selecting the best one~\cite{wong2016understanding}. It is not the optimal way to find data augmentation for specific space. It is still an open challenge to be solved. 
\subsection{\textbf{Open research questions}}
Despite the success of data augmentation techniques in different Computer Vision tasks, it still failed to solve challenges in SOTA data augmentation techniques. After thoroughly reviewing SOTA data augmentation approaches, we found several challenges and difficulties, which are yet to be solved, as it is listed below:
\begin{itemize}
    \item In image mixing techniques, label smoothing has been used. It makes sense whatever portion of images is mixed, corresponding labels should be mixed accordingly. To the best of our knowledge, none has explored label smoothing for image manipulation and image erasing subcategories - where the image part is lost. For example, if the image portion is randomly cut out in cutout data augmentation, the corresponding label should be mixed. It is an interesting open research question. 
    \item Currently, data augmentation is performed without considering the importance of an example. All examples may not be difficult for the neural network to learn, but some are. Thus, augmentation should be applied to those difficult examples by measuring the importance of the examples. How neural network behave if data augmentation is applied to those difficult examples? 
    \item In image mixing data augmentations, if we mix more than two images salient parts, that are truly participating in augmentation unlike RICAP~\cite{takahashi2018ricap}, what is its effect in terms of accuracy and robustness against adversarial attacks? Note, the corresponding labels of these images will be mixed accordingly.
    \item In random data augmentation under the auto augmentation category, the order of augmentations has not been explored. We believe it has a significant importance. What are the possible ways to explore the order of existing augmentations such as first traditional data augmentations and then image mixing or weight-based? 
    \item Finding an optimal and an ordered number of data augmentation, and the optimal number of samples to be augmented are open challenges.  For example, in randAug method, there are N optimal number of augmentations found but it is not known how many, in which order and what samples should be augmented? 
\end{itemize}
\section{Conclusion}
This survey provides a comprehensive overview of state-of-the-art (SOTA) data augmentation techniques for addressing overfitting in computer vision tasks due to limited data. A detailed taxonomy of image data augmentation approaches is presented, along with an overview of each SOTA method and the results of its application to various computer vision tasks such as image classification, object detection, and semantic segmentation. The results for both supervised and semi-supervised learning are also compiled for easy comparison purposes. In addition, the available code for each data augmentation approach is provided to facilitate result reproducibility. The difficulties and challenges of data augmentation are also discussed, along with promising open research questions that have the potential to further advance the field. This survey is expected to benefit researchers in several ways: (i) a deeper understanding of data augmentation, (ii) the ability to easily compare results, and (iii) the ability to reproduce results with available code.

\section*{Acknowledgment}
This research was supported by Science Foundation Ireland under grant numbers 18/CRT/6223 (SFI Centre for Research Training in Artificial intelligence), SFI/12/RC/2289/$P\_2$ (Insight SFI Research Centre for Data Analytics), 13/RC/2094/$P\_2$ (Lero SFI Centre for Software) and 13/RC/2106/$P\_2$ (ADAPT SFI Research Centre for AI-Driven Digital Content Technology). For the purpose of Open Access, the author has applied a CC BY public copyright licence to any Author Accepted Manuscript version arising from this submission.

\bibliographystyle{plain}
\bibliography{references.bib}

\end{document}